\documentclass{article}

\PassOptionsToPackage{numbers, compress}{natbib}

\usepackage[preprint]{neurips_2026}

\usepackage[utf8]{inputenc} 
\usepackage[T1]{fontenc}    
\usepackage{hyperref}       
\usepackage{url}            
\usepackage{booktabs}       
\usepackage{amsfonts}       
\usepackage{nicefrac}       
\usepackage{microtype}      
\usepackage[table]{xcolor}         
\usepackage{enumitem}
\usepackage{amsmath}
\usepackage{graphicx}
\usepackage{multirow}
\usepackage[export]{adjustbox}
\usepackage{subcaption}
\usepackage{pifont}          
\usepackage{makecell}        

\usepackage{minitoc}

\usepackage[nolist]{acronym}
\usepackage{soul} 

\title{Physics-IQ Verified}

\author{
Tim Rädsch\textsuperscript{ 1,2*}, 
Yuki M Asano\textsuperscript{ 3},
Hilde Kuehne\textsuperscript{ 4},
Stefan Bauer\textsuperscript{ 2,5},
\\  
~\textbf{
Priyank Jaini\textsuperscript{ 6}, 
Robert Geirhos\textsuperscript{ 6},
Carsten T.~Lüth\textsuperscript{ 1*}
}
\\\\
\textsuperscript{1}Anates Labs\\ 
\textsuperscript{2}Technical University of Munich\\ 
\textsuperscript{3}University of Technology Nuremberg\\
\textsuperscript{4}Tuebingen AI Center, University of Tuebingen\\
\textsuperscript{5}Helmholtz AI, Munich\\
\textsuperscript{6}Google DeepMind
\\\\
\texttt{research[at]anates[dot]ai}
}

\begin{document}
\doparttoc 
\faketableofcontents 
\begin{acronym}

\acro{IoU}{Intersection over Union}
\acro{MSE}{Mean Squared Error}
\acro{LLM}{Large Language Model}
\acro{VGM}{video generative model}

\acro{T2V}{text-to-video}
\acro{I2V}{image-to-video}
\acro{V2V}{video-to-video}

\end{acronym} 

\maketitle
\def\thefootnote{*}\footnotetext{Joint leads}\def\thefootnote{\arabic{footnote}}

\begin{abstract}
    \Acp{VGM} have become a new frontier that can be used not just for video generation but for a multitude of downstream tasks, including world modeling. 
To advance these tasks, a good video model must understand the physical reality of the world. 
Evaluating this understanding is an emerging field and has led to the \emph{Physics-IQ benchmark} \citep{motamed2026generative}, which quantifies this explicitly by comparing model-generated videos to real-world videos of physical experiments. 
In this work, we present a systematic audit of the \emph{Physics-IQ benchmark}, expose shortcomings and propose three solutions that sharpen how we can measure physical understanding of \acp{VGM}. Specifically, we improve prompt and ground-truth quality to reduce the influence of confounding factors and further introduce a sample-level scoring system that weights each sample and metric equally.
Our resulting benchmark, Physics-IQ Verified,
refines 57.6\% of all samples and improves over 34.8\% of prompts.
In a comparison study using six image-to-video generative models, we observe moderate but meaningful ranking changes (Kendall's $\tau = 0.46$). 
We hope Physics-IQ Verified advances the community by providing a more reliable signal toward physically accurate \acp{VGM}.
The code for the benchmark can be accessed at \href{https://github.com/google-deepmind/physics-iq-benchmark}{Physiqs-IQ Verified Github}.



\end{abstract}

\section{Introduction}
\label{s:intro}
\Acp{VGM} are increasingly positioned not merely as synthesis tools but as \emph{world models} \citep{schmidhuber1990making, lecun2022path, bruce2024genie} which simulate the physical world for complex tasks in robotics \citep{jang2025dreamgen} or as general visual task solvers~\citep{wiedemer2025video}. This use is motivated by the assumption that the next-frame prediction objective implicitly teaches the model to encode the causal structure of physical reality \citep{ha2018world}.
This framing raises an immediate question: 

\emph{How can we assess whether a model has actually learned to reason about the physical world, rather than learned to produce plausible-looking motion?}

Earlier benchmarking efforts addressed this question using distributional metrics that compare unmatched sets  of generated and real-world videos, such as Frechet Video Distance~\citep{unterthiner2019fvd} or Frechet Video Motion Distance~ \citep{liu2024fr}.
The \emph{Physics-IQ~Benchmark}~ \citep{motamed2026generative} innovated this line of work by instead comparing model-generations to ground-truth recordings from controlled real-world physical experiments instead of simulated physics environments \citep{wang2026very,bear2021physion,tung2023physion++,ates2022craft,riochet2018intphys,baradel2019cophy,yi2019clevrer,rajani2020esprit,kang2024far,bakhtin2019phyre,agarwal2025cosmos}. 
To quantify physical understanding it relies on four metrics that quantify \emph{where} action occurs, \emph{when} it occurs, \emph{how strongly} it occurs, and \emph{how closely} the generated frames match the ground truth at the pixel level.
 This design makes Physics-IQ one of the first benchmarks capable of directly measuring physical understanding rather than perceptual realism, and it has seen rapid adoption as a standard evaluation protocol for \acp{VGM} \citep{yuan2025improving,yuan2026inference,teng2025magi,OpenAI2025Sora2,zhuang2025video,liu2025bootstrapping,lu2026phys4d} also directly affecting model development ~\citep{yuan2025improving,yuan2026inference,teng2025magi, OpenAI2025Sora2,liu2025bootstrapping}.
Therefore the fidelity of its scores to actual physical understanding becomes increasingly consequential.


\begin{figure}
    \centering
    \includegraphics[width=\linewidth]{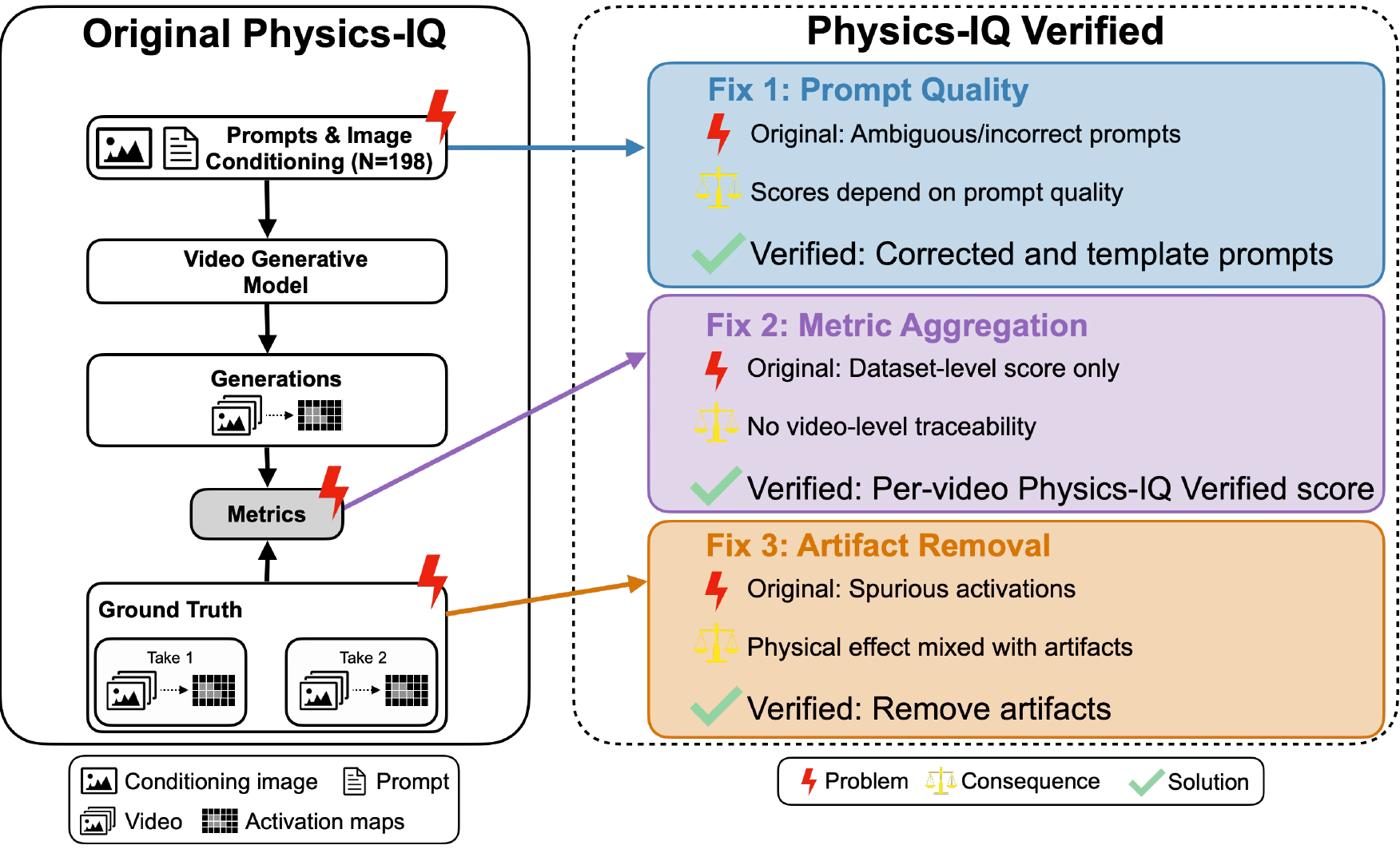}
    \caption{
    \textbf{Key improvements from the original to the verified Physics-IQ evaluation.}
    We propose three refinements to the original pipeline targeting: (1) prompt quality, (2) metric aggregation, and (3) spurious metric activations (artifacts). These improvements together sharpen the focus of the evaluation on physical understanding rather than confounding factors and also lead to a fine-grained understanding of the final score in which also all samples are weighted equally.
    We provide a detailed pipeline overview, including the original and verified metric computation, in App.~\ref{app:phys-iq-detailed-pipeline}.
    }
    \label{fig:main_figure}
\end{figure}

We present an audit of Physics-IQ proposing three distinct improvements that reduce measurement errors arising from the evaluation protocol:

\textbf{Improving Prompt quality.} Some original prompts are ambiguous in their descriptions and prompting guidelines for models are not taken into account at all. We improve the quality of unclear text prompts by providing distinctive descriptions and by adhering to model-specific best practices for prompting using a \emph{templater}. These two refinements ensure that the score reflects the capabilities of the evaluated model and minimizes the influence of suboptimal prompting. 

\textbf{Improving Metric Aggregation.} The original Physics-IQ score is only defined  on a dataset level, which also leads to samples having different influence on the final score. We define a sample-level \emph{Physics-IQ Verified} score, which allows tracing back failure modes to each individual sample and weighs all samples and metrics equally.

\textbf{Cleaning of Artifacts.} Many videos contain ``spurious metric activations'' or artifacts that are not caused by the physical phenomena. We remove these artifacts from the ground truth of the reference videos, 
leading to the score more closely measuring the physical effect rather than unrelated and possibly random events.

\begin{figure}
    \centering
    \includegraphics[width=1\linewidth]{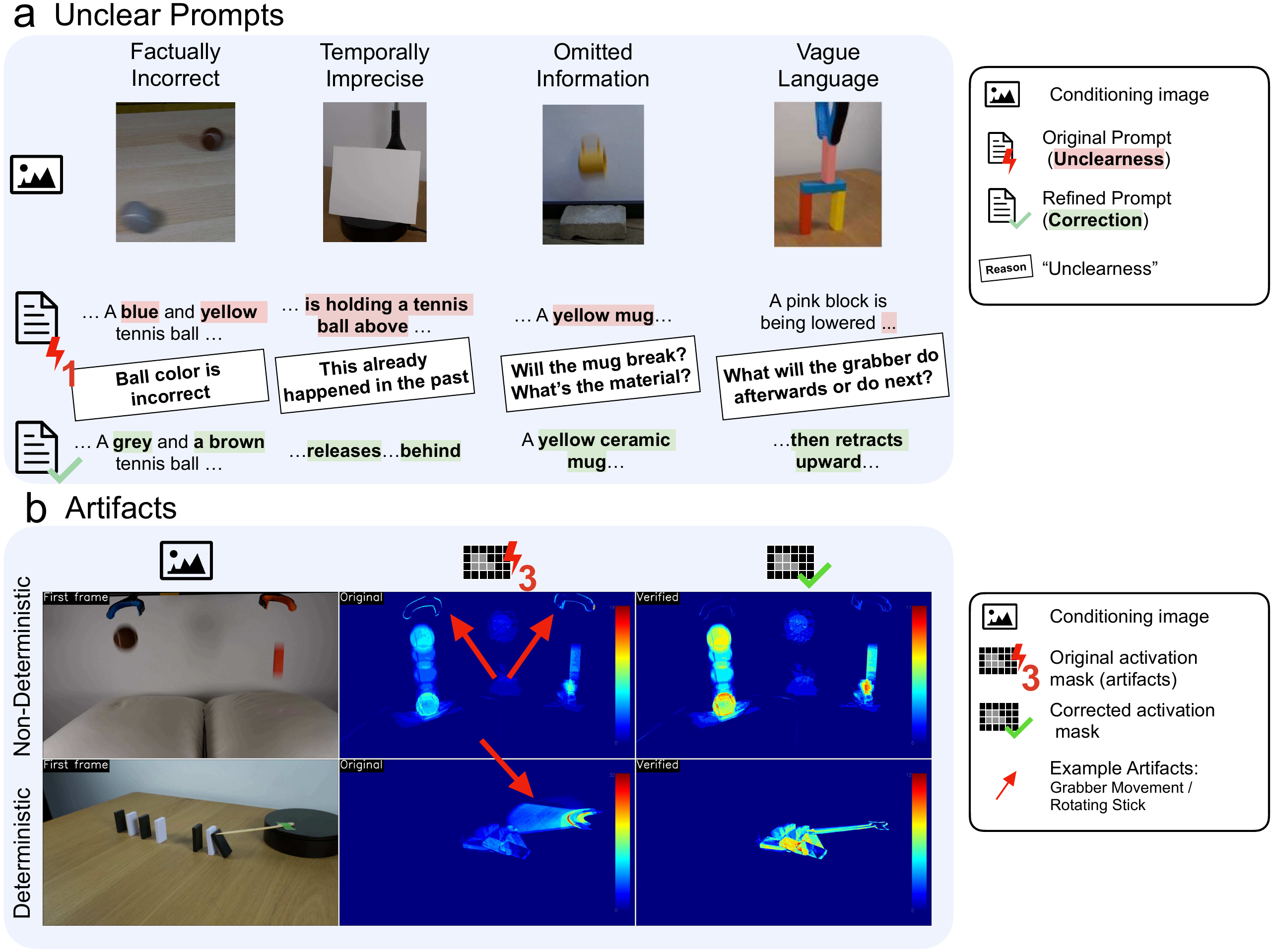}
    \caption{\textbf{Examples of unclear prompt and artifact corrections in Physics-IQ Verified.}
    \textbf{(a)}~Unclear prompts reduce the ability of either a model or human to reliably predict the physical effect as key questions with respect to the movement are not addressed. Examples for each of the four categories in decreasing order of severity from left to right alongside our corrections. 
    \textbf{(b)}~Artifacts influence the binary activations, here visualized as a temporally aggregated heatmap, arising from visual events not stemming from the physical phenomena to be observed which we categorize into non-deterministic and deterministic. 
    All three IoU-based metrics (see Sec.~\ref{s:bckg-physiq}) directly operate on these activations and compare them to activations arising from generated videos to assess whether the physical phenomena were modeled accurately.
    The occurrence of artifacts (red arrows), however, reduces the ability of these metrics to capture the physical phenomena potentially dominating the scoring as evident by the color scale in the original activations. Our cleaning directly addresses this by shifting the focus from the artifact towards the physical phenomena (here, falling ball and dominoes). More detailed examples are provided in App.~\ref{app:data_examples_improvement}.
    }
    \label{fig:prompt-artifact-examples}
\end{figure}

\begin{figure}
    \centering
    \includegraphics[width=\linewidth]{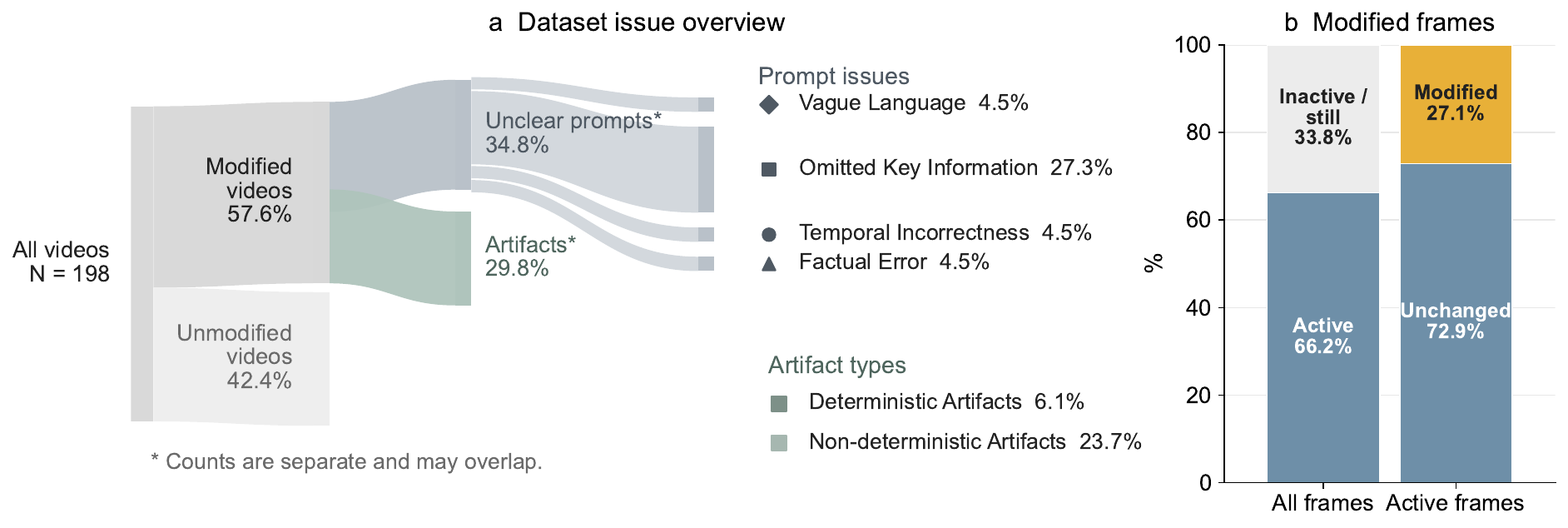}
    \caption{
    \textbf{Overview of dataset modifications and issue distributions across the 198 benchmark videos.} 
    Of the 198 videos, 69 contain unclear prompts and 59 contain artifacts, with 20 videos belonging to both groups. 
    \textbf{(a)}~Video-level overview, with flows from all videos to unclear prompts and artifacts; prompt issue categories are shown as separate counts and may overlap across videos. 
    \textbf{(b)}~Frame-level composition, showing the proportion of inactive to active frames with at least 1 activation. Within the active frames we show the proportion of unmodified to modified frames where artifacts are removed. 
        }
    \label{fig:actifact_screening}
\end{figure}

Taken together, these contributions constitute \emph{Physics-IQ Verified}: a refined benchmark that more faithfully reflects the ability of \acp{VGM} to model physical phenomena and allows a more fine-grained analysis of results by tracing back scores to a sample level. 
We provide an overview of our improvements in Figure~\ref{fig:main_figure} that highlights where the original benchmark is improved.
The refinement removes possible measurement errors in 57.6\% of all samples, influencing 29.8\% of videos, correcting over 34.8\% of prompts that are highly ambiguous (examples Figure~\ref{fig:prompt-artifact-examples} and detailed statistics Figure~\ref{fig:actifact_screening}), while it also provides a template-based prompt structure with more accurate descriptions for all videos visualized in Figure~\ref{fig:prompt-example}.
The benchmark is hosted at: \url{https://github.com/google-deepmind/physics-iq-benchmark}

Our evaluation of six \ac{I2V} \acp{VGM} using both the original and verified evaluation finds that models react differently to the improvements in evaluation, which leads to the overall ranking of models changing substantially.
This highlights that \ac{VGM} benchmarks must be carefully designed, so that models are tested on the effect of interest and to the best of their ability. 



\section{Background: The original Physics-IQ benchmark}
\label{s:bckg-physiq}
The Physics-IQ benchmark contains 66 distinct physical experiments covering solid dynamics, fluid dynamics, thermodynamics, optics and magnetism.
Each experiment is captured from three viewing angles and carried out twice resulting in overall $66 \times 3 \times 2=396$ videos (referred to as GT1 and GT2). These 8 second videos are then split into a 3 second conditioning part, and a 5 second ``ground truth'' video continuation for comparison. Each scenario includes an additional text description for 
conditioning. For the first 198 videos (ID001--198), switch frames mark the 
exact 3-second point where generation for the video generative model should begin.
These switch frames, alongside previous video frames, can also be used as conditioning input for image-to-video or video-to-video models. The generation of \acp{VGM} is therefore constrained to 5 second videos on this first set of videos.
The second set of videos (ID199--396) consists of second takes.
These takes are used to compute the physical variation between 
identical setups.
This variation serves as an upper performance ceiling, representing 
natural trial-to-trial variability.

Each video presents a physical experiment in which observable phenomena unfold after the switch-frame. The model's task is to predict these phenomena based on a full prompt, whose composition depends on the model type: a text prompt alone for \ac{T2V} models, an image combined with text for \ac{I2V} models, or a video clip or multiframe input combined with text for \ac{V2V} models.

Performance is measured using four metrics designed to quantify 
how closely the generated output replicates the physical phenomena. Three are activation-based\citep[Algo. 2]{motamed2026generative} Intersection over Union (IoU) metrics, 
and one is a pixel-based Mean Squared Error (MSE) metric:
1)~\emph{Spatial IoU}: Where does action happen?
2)~\emph{Spatiotemporal IoU}: Where \& when does action happen?
3)~\emph{Weighted spatial IoU}: Where \& how much does action happen?
4)~\emph{\ac{MSE}}: How does action happen?

To compute the final Physics-IQ score, metric values are averaged 
and divided by the physical variation. This is followed by a weighted summation, with a negative sign 
applied to the MSE.
The physical variation is obtained by computing the mean value for each of these metrics in the same way as for a normal evaluation but using the first and second take for each experiment. We give a detailed description of the used metrics with a clear mathematical notation in App.~\ref{app:metric-discussion}.

\textbf{Physics-IQ's position among other benchmarks.}
Unlike judgment-based benchmarks that assess whether a video appears physically plausible~\citep{bansal2024videophy,meng2024towards}, or simulation benchmarks using synthetic data that test predefined physical rules~\citep{wang2026very, bear2021physion,tung2023physion++,ates2022craft,riochet2018intphys,baradel2019cophy,yi2019clevrer,rajani2020esprit,kang2024far,bakhtin2019phyre}, Physics-IQ~\citep{motamed2026generative} compares generated continuations to real-world recordings of the same physical setup. 
This reference-based design makes Physics-IQ especially valuable because it provides a concrete physical target rather than a categorical plausibility judgment; at the same time, it makes the benchmark particularly sensitive to the quality of the ground-truth recordings. If prompts, reference activations, or aggregation choices include confounding factors, they directly change what physical effect is treated as the measurement target, motivating our audit. 
We provide a more detailed comparison to related benchmark families in App.~\ref{app:relworks}.
\section{Physics-IQ Verified: Sharpening How Physical Understanding is Assessed}
\label{s:improvements}


\subsection{Improving text prompts}
\label{ss:prompt}

\acp{VGM} are steered through visual prompts, including conditioning frames of the initial state and text prompts describing the physical process. The prompt quality directly bounds what the benchmark can measure. Our proposed improvements address two sources of measurement error in the original benchmark: unclear prompts, and a lack of proper structure for specific models. We address each in turn, starting with a definition of a well designed prompt.


\emph{A well-designed prompt for assessing VGMs' ability to model physics is a text description, accompanied by a conditioning frame or video, that clearly specifies the full experimental setup and the catalyst of the physical phenomenon, without revealing how that phenomenon unfolds.}

The prompt should function as an exam question: a human provided with the prompt and start frame should be able to predict the experimental outcome with high confidence, yet the prompt must not make the answer obvious, lest it trivialize the generation task. Any ambiguity left unresolved by the prompt introduces degrees of freedom in the output that are orthogonal to physical understanding and therefore inflates metric variance or reduces performance irreducibly. Therefore, we depart from the original benchmark's focus on scene description (\citet[p.~2]{motamed2026generative}) in favor of clear experimental instructions. 

\begin{figure}
    \centering
    \includegraphics[width=\linewidth]{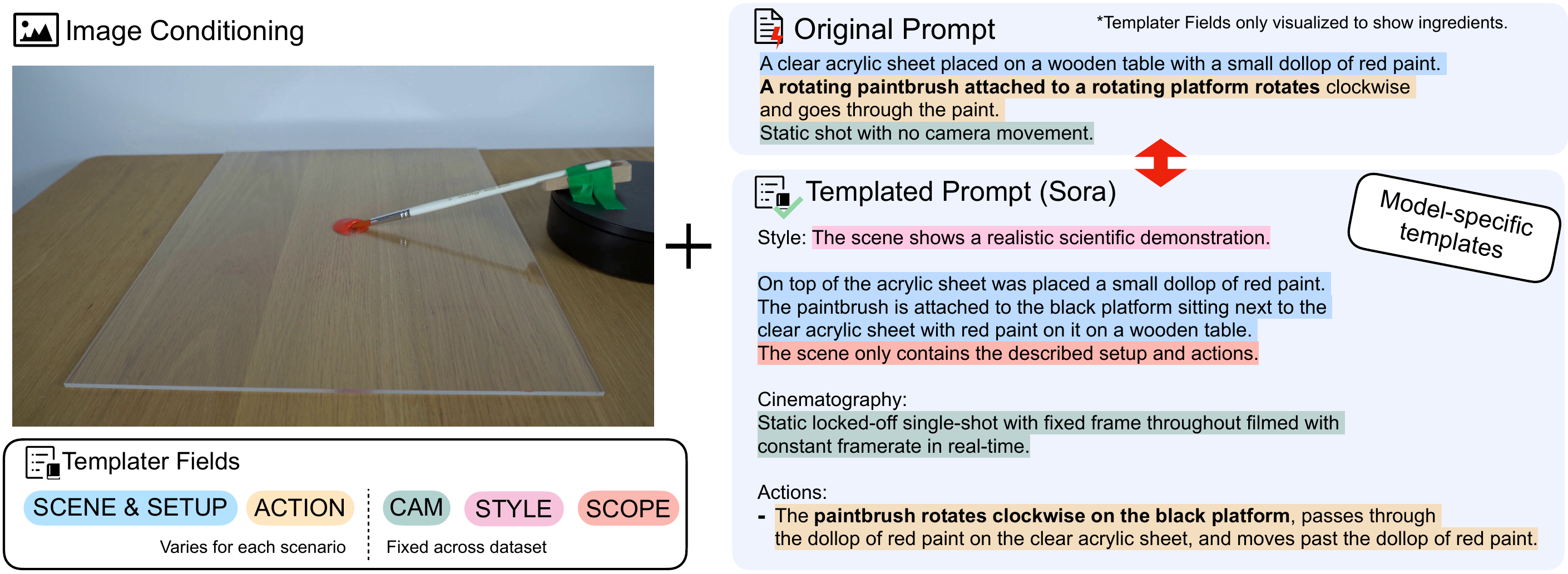}
    \caption{\textbf{Full prompt improvement showcasing correction and templater.}
    The original prompt does not adhere to the best-practices of the model providers. We address this by grouping the information contained in a prompt into six fields (each color denoting a separate field where \texttt{SETUP} \& \texttt{SCENE} are merged for this cases). These fields can be used by custom templaters for each model, here visualized for Sora. 
    The \texttt{ACTION} field contains the experiment description, the \text{CAM} field now contains more explicit descriptions of the video format, the \texttt{STYLE} field ensures that the model is aware that scientific experiments are conducted, and the \texttt{SCOPE} field ensures that the model is aware that it should not hallucinate new interactions. The latter two fields are new additions.
    Finally, in this specific example the action is also factually incorrect (\textbf{bold text}) stating that the paintbrush rotates on a rotating platform, in fact it rotates on the platform.
    }
    \label{fig:prompt-example}
\end{figure}
We provide one concrete example with all the resulting changes detailed in the rest of this section in Figure~\ref{fig:prompt-example}.

\subsubsection{Clarifying Unclear Prompts}

Unclear prompts fail to narrow the space of plausible scenarios towards the specific scenario observed in the physical experiment. We identify four severity levels, ranging from making correct generation impossible to merely increasing output variance (see Figure~\ref{fig:prompt-artifact-examples} for examples). These are, in order of severity: 

(1)~\emph{Factually~incorrect}: does not match what happens in the video;
(2)~\emph{Temporally~imprecise}: fails to distinguish actions that have already occurred prior to the conditioning frame from following actions that should be generated;
(3)~\emph{Omitted~key~information}: lacks information necessary to accurately model the physical effect;
(4)~\emph{Vague~language}: describes the observed action in terms that are too imprecise to sufficiently constrain the generation.

Factual incorrectness and temporal impreciseness make accurate generation impossible in principle.
Omitted key information and vague language increase output variance by leaving physical degrees of freedom unconstrained.
Each of these reduce the ability of both \acp{VGM} and humans to predict the physical effect reliably. This can bias the final score to reflect prompt clarity rather than model capability. Thus, here, we carefully screened and applied minimally invasive corrections yielding a complete set of \emph{updated descriptions}.

\subsubsection{Adhering to the Prompt--Model Interface}
Independent of content quality, the original prompts are not structured according to the input conventions of the \acp{VGM} being evaluated.
This manifests in the generation being insufficiently conditioned on the text prompt.
Since the benchmark's goal is to assess physical reasoning rather than robustness to naive user inputs, prompts should simulate an experienced user familiar with the target model. 

To ensure consistent conditioning, we decompose each prompt into six structured fields. These six fields are used by model-specific templaters, which create the text prompt according to providers' best practices (an example is shown in Figure~\ref{fig:prompt-example}). 
Three fields, namely \texttt{SETUP}, \texttt{SCENE}, and \texttt{ACTION}, capture variable, scenario-specific information adapted from the original prompts, while \texttt{CAM}, \texttt{STYLE}, and \texttt{SCOPE} remain consistent across all 66 scenarios. The latter two fields represent novel additions absent from the originals, each targeting a systematic gap. 

\texttt{STYLE} constrains the rendering register to ``...a realistic scientific demonstration'', preventing stylised or cartoonish outputs. \texttt{SCOPE} instructs ``only contains the described setup and actions'' to ensure the model is aware that no new actors or interactions enter the scene, suppressing hallucinated intrusions. The \texttt{CAM} field is changed to use descriptive cinematographic language describing the  expected video in detail to ensure that it is sufficiently clear: \emph{``Static locked-off single-shot with fixed frame throughout, filmed at constant framerate in real-time.''}. The importance of camera guidance is also evident in other works when evaluating \acp{VGM} \citep{wiedemer2025video}.

A core principle while rewriting the prompts into our six fields is to express all instructions in positive terms as text-based negations are poorly handled by many models \citep{truong2023language,garcia2023not,parcalabescu2022valse,alhamoud2025vision,conwell2024relations} and some model providers explicitly discourage them.\footnote
{
e.g. FLUX: \url{https://docs.bfl.ml/guides/prompting_summary}}
We provide more details with respect to this rewrite, the templater and the design process in App.~\ref{app:templater}.

\subsection{Improving Aggregation: Enforcing Equal Weights for each Sample and Metric}
\label{ss:unbounded_scores}
The original Physics-IQ score aggregates metrics across the entire dataset of $N$ samples as follows:
\begin{equation}
s^{\text{Physics-IQ}} = c_{[0,1]} \left( \frac{1}{3} 
\sum_{M_{\text{IoU}} \in \{\text{SP,ST,WS}\}} 
\frac{\sum_{n=1}^N v^{M_{\text{IoU}}}_n}{\sum_{n=1}^N r^{M_{\text{IoU}}}_n} 
- \frac{\sum_{n=1}^N (v^{\text{MSE}}_n - r^{\text{MSE}}_n)}{N} \right)
\label{eq:physiq_original}
\end{equation}
Here, $v^{\text{M}}_{n}$ is the metric value comparing the generated video to the reference (GT 1) for sample $n$ for the four metrics: spatial (SP)-, spatiotemporal (ST)-, weighted spatial (WS)-IoU and \ac{MSE}. The clipping operation $\operatorname{c}_{[0,1]}$ ensures the final score remains 
within $[0, 1]$. The physical variation $r^M_n$ acts as a 
normalization factor. It is obtained by comparing the second take 
of an experiment (GT 2) to the first take (GT 1), treating GT 2 as a baseline generation.

This dataset-wide aggregation has two structural issues, both 
stemming from the summation inside the denominator. First, the physical variation $r$ should reflect an upper bound for each specific experiment's score. Averaging $r$ across the dataset invalidates this upper bound. Consequently, experiments with low physical 
variation are down weighted because they can never reach a score of 1.
Conversely, experiments with high physical variation are up weighted, 
as their individual scores can exceed 1.
Second, dataset-wide 
calculation obscures sample-level failures, making it difficult to 
trace low benchmark scores to specific failure modes. This reduces 
the benchmark's utility for steering model development.

To solve these issues, we define the \emph{Physics-IQ Verified score} directly at the sample level ($n$). We aggregate the subscores 
using the arithmetic mean so that improvements in any metric are clearly 
reflected in the sample's total score. To ensure the MSE is 
interpreted similarly to the IoU metrics (where higher is better), 
we define its influence as the inverse ratio of the MSE physical 
variation. This yields the following per-sample score:
\begin{equation}
s^{\text{Physics-IQ Verified}}_n = \frac{1}{4} \left( c_{[0,1]} 
\left( \frac{r^{\text{MSE}}_n}{v^{\text{MSE}}_n} \right) + 
\sum_{M_{\text{IoU}} \in \{\text{SP, ST, WS}\}} c_{[0,1]} 
\left( \frac{v^{M_{\text{IoU}}}_n}{r^{M_{\text{IoU}}}_n} \right) \right)
\label{eq:physics-iq-verified}
\end{equation}

The final Physics-IQ Verified score is the arithmetic mean across 
all samples: $s^{\text{Physics-IQ Verified}} = \frac{1}{N} \sum_{n=1}^N 
s^{\text{Physics-IQ Verified}}_n$.
Further details regarding the 
computation and the drawbacks of the original score are provided in App.~\ref{app:metric-discussion}.

\subsection{Cleaning of Spurious Metric Activations or Artifacts}
\label{ss:artifacts}
All three IoU-based metrics used in Physics-IQ operate on activation maps\citep[Algo. 2]{motamed2026generative} that are derived from the visual differences of neighboring video frames, for both ground truth videos and generated videos. 
Because this applies to both ground truth and generated videos, 
the quality of the ground truth activations is crucial.
High-quality activations ensure the metrics assess physical phenomena 
rather than ``spurious activations'' or artifacts. 
We, therefore, define:

\emph{An artifact as a metric activation caused by a visual event that is not part of the physical effect under observation. We distinguish them into two subtypes based on predictability}:
\begin{itemize}[nosep, leftmargin=*]
    \item \emph{Deterministic artifacts} stem from events that are specifiable from the prompt or experimental setup (e.g., a rotating apparatus). They are in principle predictable, but generate activation signal that is attributable to the apparatus rather than the physical phenomenon of interest.
    \item \emph{Non-deterministic artifacts} arise by chance during recording and are absent from any prompt or experimental specification.
\end{itemize}
We show examples for both deterministic and non-deterministic artifacts alongside the result of our corrections in Figure~\ref{fig:prompt-artifact-examples} and provide their prevalence in Figure~\ref{fig:actifact_screening}.

Both artifact types hinder assessment of physical understanding, but through distinct mechanisms. Deterministic artifacts bias the metric by adding activation signal that reflects apparatus behaviour rather than physical understanding, biasing scores in a structured way. 
Non-deterministic artifacts are more damaging from a measurement 
perspective.
Because they are neither prompt-specified nor experimentally controlled, 
no model or human can anticipate them. 
This contributes entirely irreducible variance or bias to the 
benchmark scores. 

We address both artifact types with a targeted removal strategy using 
manual annotations of the ground truth videos.
First, we use \texttt{end\_effect\_frame}s to indicate when the physical 
phenomenon ends, removing any artifacts that occur afterwards.
Second, we use \texttt{freeze\_area}s to pinpoint the spatial location 
and timing of artifacts occurring \emph{during} the physical phenomenon.
This allows us to remove artifacts that happen before the 
\texttt{end\_effect\_frame}.  Details about artifact removal are provided in App.~\ref{app:data_examples_improvement}.

\section{Experiments}
\label{s:exp}

\begin{figure}
    \centering
    \includegraphics[width=\linewidth]{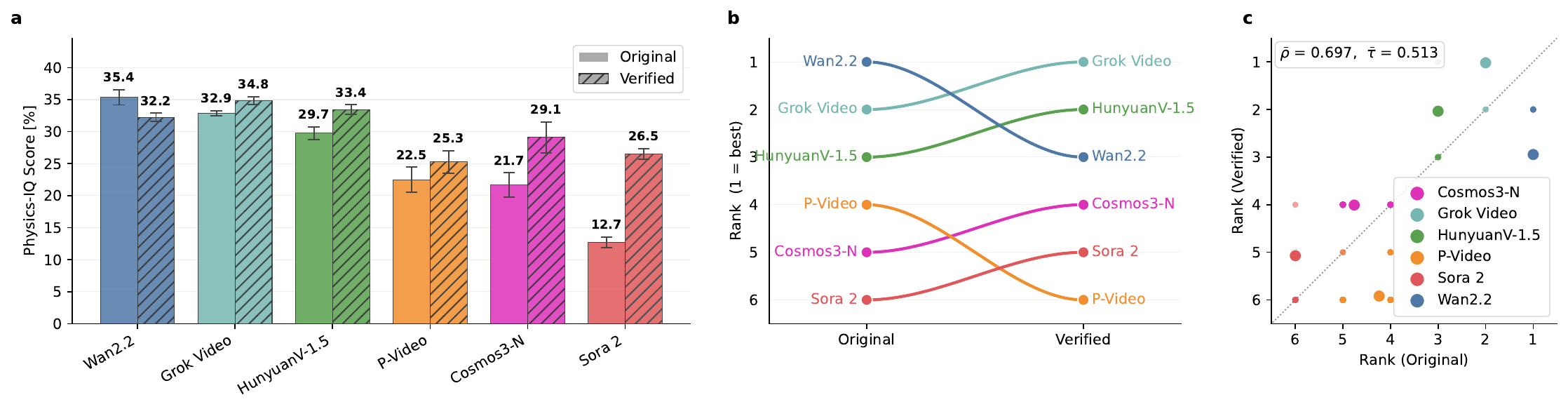}
    \caption{\textbf{Comparison of Physics-IQ scores in its original and our proposed verified form.} 
    \textbf{(a)}~Side-by-side comparison of final Physics-IQ scores for each model. For all models, with the exception of Wan 2.2, the scores increase for the verified evaluation. Sora 2 shows the largest increase in scores. 
    T-denotes the standard deviations across four different runs.
    \textbf{(b)}~Ranking bump plot highlighting the differences in ranking with Wan 2.2 moving from first to third and Sora 2 jumping from sixth to fifth place, while Cosmos3-N moves from fifth to fourth.
    \textbf{(c)}~Bootstrap analysis ranking scatter plot. Large dots indicate the mean rank, while the smaller faint dots indicate the frequency with stronger color indicating more frequent ranks. Both the mean Spearman-$\rho$ and Kendall-$\tau$ signal meaningful ranking differences. 
    }
    \label{fig:original-verified}
\end{figure}

\paragraph{Experimental Setup.}
We evaluate six \ac{I2V} \acp{VGM}: three open-source, \emph{Wan 2.2} \citep{wan2025wan}, \emph{HunyuanV-1.5} \citep{kong2024hunyuanvideo} and \emph{Cosmos3-N}~\citep{agarwal2026cosmos}, and three closed-source, \emph{Sora 2} (v2025-10) \citep{OpenAI2025Sora2}, \emph{P-Video} \citep{pruna} and \emph{Grok} Imagine \emph{Video} \citep{xai2026grokimagineapi}. 
We provide details with respect to all models in Table~\ref{tab:model_generation_settings}.
Each model generates four complete sets of videos on the Physics-IQ dataset for both the original prompts (\emph{op}) and our best-practice prompts (\emph{bpp}), where each set consists of 198 videos following the standard \ac{I2V}-protocol \citep{motamed2026generative}.
We perform evaluations in a factorial design that isolates the influence of each of our proposed evaluation improvements leading to 8 settings: $\times 2$~Prompt~(op~\&~bpp)~$\times 2$~GT~(original~\&~verified)~$\times 2$~score~(original~\&~verified).
Detailed results are provided in App.~\ref{app:sec_additional_results}.

\paragraph{Method of Analysis.}
The resulting rankings are analyzed using Kendall's-$\tau$ \citep{kendall1945treatment} and Spearman's-$\rho$ \citep{spearman1961proof}; both metrics range from -1 to 1 and larger values indicate more agreement between rankings. 
Additionally we perform bootstrap analysis where 500 complete sets of videos of size 198 are generated by drawing for each video id the corresponding video from one of the four original sets. Based on this we estimate mean and 95\% confidence intervals for Spearman's-$\rho$ and Kendall's-$\tau$.
We analyze absolute changes using Cohen's d \citep{cohen1977statistical} to estimate the influence and Wilcoxon tests \citep{wilcoxon1945individual} to confirm statistical significance. 
During testing we evaluate each model run as an independent event following \citet{demvsar2006statistical}.

\subsection{Comparing Original and Verified Evaluation}
\label{ss:ranking}
We compare the results of the original and our proposed verified evaluation using both the artifact removed ground truth and the best-practice prompts in Figures~\ref{fig:original-verified}a\&b.
Overall, the scores increase for most models using Physics-IQ Verified compared to the original, mostly stemming from the improved prompts and our verified scores yielding higher values. 
Sora 2\footnote
{The Sora 2 performance is notably worse in April 2026 than in October 2025. We confirm this in App. Tables~\ref{tab:sora-sanity-original_score}\&\ref{tab:sora-sanity-verified_score}.}
and Cosmos-3N have the highest increase in performance both outperforming P-Video. Overall, the verified evaluation produces a rank reshuffling: Grok Video and HunyuanV-1.5 move ahead of Wan 2.2 (the only model to reduce the score), Cosmos3-N and Sora 2 improve their positions, while P-Video falls from fourth to last place.
The Spearman ($\rho=0.65$) and Kendall ($\tau=0.46$) correlations between rankings indicate moderate but meaningful changes. This is corroborated by the bootstrap analysis in Figure~\ref{fig:original-verified}c: within-ranking correlations exceed $0.9$, and their 95\% confidence intervals do not overlap with the cross-evaluation ranking correlations ($\bar{\rho}=0.697$, $\bar{\tau}=0.513$; see App.\ Figure~\ref{fig:bootrank-analysis} for details).

As these changes in scores and ranking are the result of three separate changes, we trace back the influence for each of these changes by assessing their impact on the original evaluation. We will start by giving the high-level takeaways and then discuss the details following this.

Overall better prompts improve the quality for all models with Sora 2 benefiting from it the most with Wan 2.2 being the only exception losing performance. Meanwhile artifact removal decreases the scores for all models but again most notably for Wan 2.2 indicating that some of its better score over other models stems from confounding effects. 
Changing the score from the original formulation to our sample level score yielded no change in overall ranking but increased the Physics-IQ score for all models.


\subsection{Systematic Impact Assessment of each Improvement on the Original Evaluation}
\begin{figure}
    \centering
    \begin{subfigure}{0.49\linewidth}
        \includegraphics[width=1\linewidth]{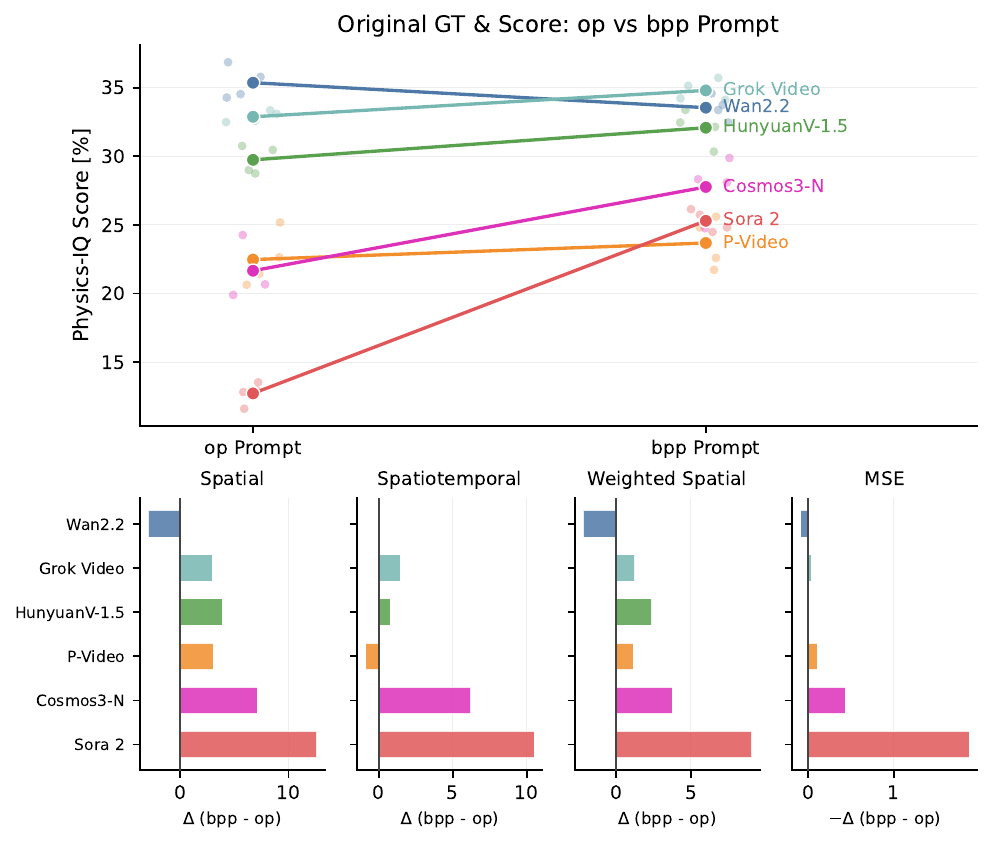}    
        \caption{Improved Prompts}
        \label{fig:prompt-analysis}
    \end{subfigure}
    \begin{subfigure}{0.49\linewidth}
        \includegraphics[width=1\linewidth]{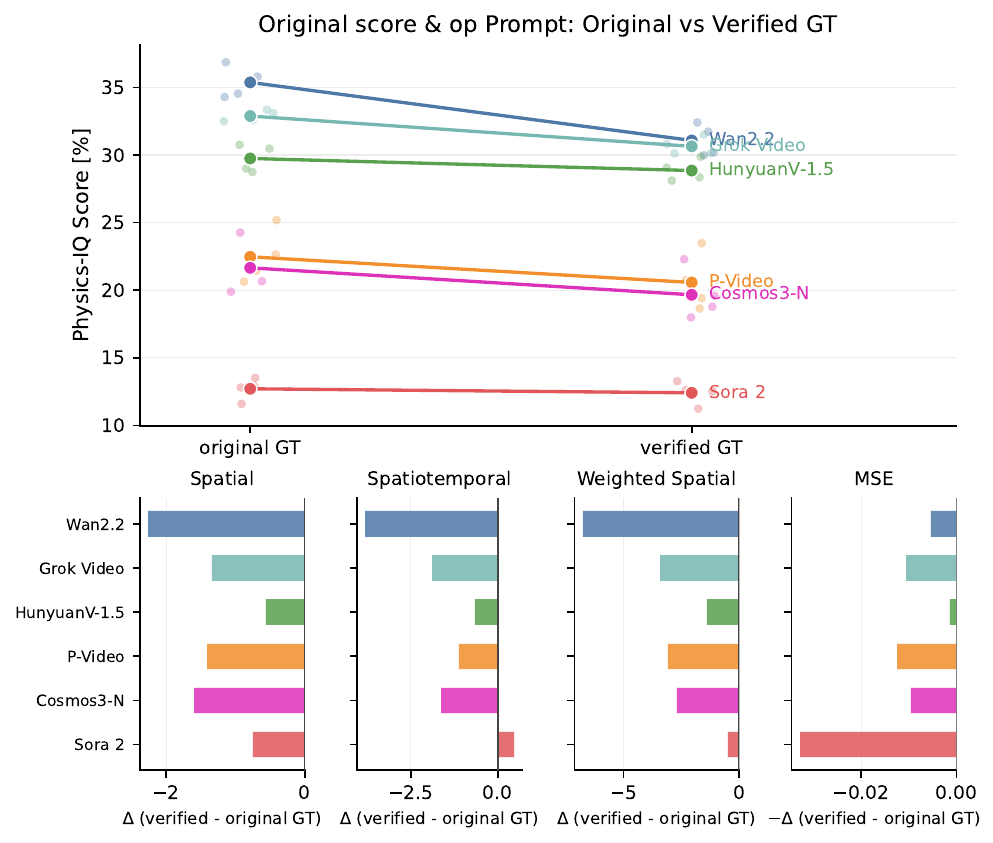}
        \caption{Artifact Cleaning}
        \label{fig:eval-analysis}
    \end{subfigure}
    \caption{\textbf{The Influence of Prompts and Artifacts on the resulting scores.}
    \textbf{(a)} Prompts: All models with the exception of Wan 2.2 benefit from the inclusion of the best-practice prompts (bpp) over original prompts (op). Wan 2.2 is the only model for which the performance decreases.
    \textbf{(b)} Artifacts: Here denoted as original GT (with artifacts) and verified GT (without artifacts). All models show a reduction in absolute performance when assessed with the verified evaluation with reductions being overall largest for the weighted spatial score. Wan 2.2 is subject to the largest absolute performance reduction. 
    }
    \label{fig:part-analysis}
\end{figure}

\paragraph{Influence of prompts.}
Best-practice prompts (bpp) yield significantly significantly better sub-scores than original prompts (op) across all primary metrics in the original evaluation
(Wilcoxon signed-rank: all $p <0.05$), with medium-to-large effect sizes (Cohen's $d\geq0.55$ for all scores), as shown in Figure~\ref{fig:prompt-analysis}. The magnitude of improvement is model-dependent: for Sora 2, bpp prompts substantially reduce unwanted camera motion present under op prompts, driving large gains across all metrics. Wan 2.2 is the only model for which performance decreases under bpp, despite following guidelines for prompts.

\paragraph{Investigating influence of artifacts.}
Removing evaluation artifacts significantly reduces performance across all IoU-based metric scores and the original Physics-IQ score (Wilcoxon signed-rank: all $p \ll 10^{-5}$), with large effect sizes (Cohen's $d \leq -1$ for all scores), as visualized in Figure~\ref{fig:eval-analysis}. To identify the source of these reductions, we decompose score changes into numerator and denominator contributions. For most metrics, physical variance is nearly identical across protocols, so the score reduction is attributable entirely to the numerator. For the spatiotemporal metric, physical variance increases by ${\approx}17\%$ under the verified protocol, introducing a denominator effect that mechanically suppresses scores independently of model behavior. These two mechanisms are structurally distinct and scores should not be compared across protocols without normalizing for this variance difference.

We hypothesize that the high degrees of freedom in the original prompts make it unlikely that any model interprets the intended physical scenario consistently, which would explain why bpp prompts reduce score variance in addition to improving mean performance; however, we did not observe this.
We suspect that this might stem from the large degree of freedom in scenarios described by the original prompt making it very unlikely that the model is interpreting the prompt by chance close enough to its intended purpose to increase scores.

\paragraph{Influence or Benefit of Proposed Score on the Ranking.} In our evaluation our proposed score does yield higher Physics-IQ scores than the original formulation for all models.
This almost uniform increase in scores does not change the ranking which is confirmed by the bootstrap ranking analysis resulting in almost perfect alignment values $\approx1$ for both $\bar{\rho}$ and $\bar{\tau}$.
Details on both evaluations are provided in  App. Figures~\ref{fig:score-comparison}\&\ref{fig:score-bootstrap}.
In our evaluation the main benefit of the proposed score therefore lies in the improved granularity which allows to trace back the influence of individual samples on the final score.

\section{Conclusion}
\label{s:conclusion}
We presented a systematic audit of the influential Physics-IQ benchmark \citep{motamed2026generative}, whose finding that visual realism and physical understanding are largely uncorrelated has shaped subsequent work in the field. 
In our assessment, we identify three sources of measurement error and propose targeted solutions for each: text prompt improvements, artifact removal, and sample-wise score aggregation. 
Our experiments using six \acp{VGM} confirm that these changes impact the final evaluation in a significant way, with artifact removal reducing and improved prompts increasing absolute scores. Together, these refinements also change the final ranking of models. 
By providing the improved \emph{Physics-IQ Verified} benchmark, we improve the measurement of physics of \acp{VGM} and hope to enable building the next generation of physically accurate \acp{VGM}.

\begin{ack}
The authors would especially like to thank Tassilo Wald for his detailed feedback on multiple drafts of this paper. We also thank Pruna AI for providing model credits to access their model.
P.J. and R.G. contributed in an advisory capacity. 
\end{ack}

{
\small
\bibliography{bibliography,bibliography_related}

@article{wilcoxon1945individual,
  title={Individual comparisons by ranking methods},
  author={Wilcoxon, Frank},
  journal={Biometrics bulletin},
  volume={1},
  number={6},
  pages={80--83},
  year={1945},
  publisher={JSTOR}
}

@book{cohen1977statistical,
  title={Statistical power analysis for the behavioral sciences, Rev},
  author={Cohen, Jacob},
  year={1977},
  publisher={Lawrence Erlbaum Associates, Inc}
}

@article{meng2024towards,
  title={Towards world simulator: Crafting physical commonsense-based benchmark for video generation},
  author={Meng, Fanqing and Liao, Jiaqi and Tan, Xinyu and Shao, Wenqi and Lu, Quanfeng and Zhang, Kaipeng and Cheng, Yu and Li, Dianqi and Qiao, Yu and Luo, Ping},
  journal={arXiv preprint arXiv:2410.05363},
  year={2024}
}

@misc{xai2026grokimagineapi,
  author       = {{xAI}},
  title        = {Grok {Imagine} {API}: State-of-the-Art Video Generation across Quality, Cost, and Latency},
  year         = {2026},
  howpublished = {\url{https://x.ai/news/grok-imagine-api}},
  note         = {Accessed: 2026-04-29}
}

@misc{pruna,
    title = {Efficient Machine Learning with Pruna},
    year = {2023},
    note = {Software available from pruna.ai, Accessed: 2026-04-29},
    url={https://www.pruna.ai/}
}

@misc{OpenAI2025Sora2, 
    title={Sora 2 system card openai September 30, 2025 1}, 
    url={https://cdn.openai.com/pdf/50d5973c-c4ff-4c2d-986f-c72b5d0ff069/sora_2_system_card.pdf}, 
    journal={Sora 2 System Card, Accessed: 2026-04-29}, 
    author={Open AI}, 
    year={2025}, 
    month={Sep}
}

@article{kong2024hunyuanvideo,
  title={Hunyuanvideo: A systematic framework for large video generative models},
  author={Kong, Weijie and Tian, Qi and Zhang, Zijian and Min, Rox and Dai, Zuozhuo and Zhou, Jin and Xiong, Jiangfeng and Li, Xin and Wu, Bo and Zhang, Jianwei and others},
  journal={arXiv preprint arXiv:2412.03603, Accessed: 2026-04-29},
  year={2024}
}

@article{wan2025wan,
  title={Wan: Open and advanced large-scale video generative models},
  author={Wan, Team and Wang, Ang and Ai, Baole and Wen, Bin and Mao, Chaojie and Xie, Chen-Wei and Chen, Di and Yu, Feiwu and Zhao, Haiming and Yang, Jianxiao and others},
  journal={arXiv preprint arXiv:2503.20314, Accessed: 2026-04-29},
  year={2025}
}

@article{spearman1961proof,
  title={The proof and measurement of association between two things.},
  author={Spearman, Charles},
  year={1961},
  publisher={Appleton-Century-Crofts}
}

@article{kendall1945treatment,
  title={The treatment of ties in ranking problems},
  author={Kendall, Maurice G},
  journal={Biometrika},
  volume={33},
  number={3},
  pages={239--251},
  year={1945},
  publisher={JSTOR}
}

@article{yuan2026inference,
  title={Inference-time Physics Alignment of Video Generative Models with Latent World Models},
  author={Yuan, Jianhao and Zhang, Xiaofeng and Friedrich, Felix and Beltran-Velez, Nicolas and Hall, Melissa and Askari-Hemmat, Reyhane and Han, Xiaochuang and Ballas, Nicolas and Drozdzal, Michal and Romero-Soriano, Adriana},
  journal={arXiv preprint arXiv:2601.10553},
  year={2026}
}

@article{demvsar2006statistical,
  title={Statistical comparisons of classifiers over multiple data sets},
  author={Dem{\v{s}}ar, Janez},
  journal={Journal of Machine learning research},
  volume={7},
  number={Jan},
  pages={1--30},
  year={2006}
}

@article{zhang2025morpheus,
  title={Morpheus: Benchmarking physical reasoning of video generative models with real physical experiments},
  author={Zhang, Chenyu and Cherniavskii, Daniil and Tragoudaras, Antonios and Vozikis, Antonios and Nijdam, Thijmen and Prinzhorn, Derck WE and Bodracska, Mark and Sebe, Nicu and Zadaianchuk, Andrii and Gavves, Efstratios},
  journal={arXiv preprint arXiv:2504.02918},
  year={2025}
}

@article{yuan2025improving,
  title={Improving the Physics of Video Generation with VJEPA-2 Reward Signal},
  author={Yuan, Jianhao and Zhang, Xiaofeng and Friedrich, Felix and Beltran-Velez, Nicolas and Hall, Melissa and Askari-Hemmat, Reyhane and Han, Xiaochuang and Ballas, Nicolas and Drozdzal, Michal and Romero-Soriano, Adriana},
  journal={arXiv preprint arXiv:2510.21840},
  year={2025}
}

@article{ha2018world,
  title={World models},
  author={Ha, David and Schmidhuber, J{\"u}rgen},
  journal={arXiv preprint arXiv:1803.10122},
  volume={2},
  number={3},
  pages={440},
  year={2018}
}

@article{liu2024fr,
  title={Fr$\backslash$'echet Video Motion Distance: A Metric for Evaluating Motion Consistency in Videos},
  author={Liu, Jiahe and Qu, Youran and Yan, Qi and Zeng, Xiaohui and Wang, Lele and Liao, Renjie},
  journal={arXiv preprint arXiv:2407.16124},
  year={2024}
}

@misc{unterthiner2019fvd,
title={{FVD}: A new Metric for Video Generation},
author={Thomas Unterthiner and Sjoerd van Steenkiste and Karol Kurach and Rapha{\"e}l Marinier and Marcin Michalski and Sylvain Gelly},
year={2019},
url={https://openreview.net/forum?id=rylgEULtdN}
}

@inproceedings{jang2025dreamgen,
  title={DreamGen: Unlocking Generalization in Robot Learning through Video World Models},
  author={Jang, Joel and Ye, Seonghyeon and Lin, Zongyu and Xiang, Jiannan and Bjorck, Johan and Fang, Yu and Hu, Fengyuan and Huang, Spencer and Kundalia, Kaushil and Lin, Yen-Chen and others},
  booktitle={Conference on Robot Learning},
  pages={5170--5194},
  year={2025},
  organization={PMLR}
}

@inproceedings{bruce2024genie,
  title={Genie: Generative interactive environments},
  author={Bruce, Jake and Dennis, Michael D and Edwards, Ashley and Parker-Holder, Jack and Shi, Yuge and Hughes, Edward and Lai, Matthew and Mavalankar, Aditi and Steigerwald, Richie and Apps, Chris and others},
  booktitle={Forty-first International Conference on Machine Learning},
  year={2024}
}

@article{lecun2022path,
  title={A path towards autonomous machine intelligence version 0.9. 2, 2022-06-27},
  author={LeCun, Yann and others},
  journal={Open Review},
  volume={62},
  number={1},
  pages={1--62},
  year={2022}
}

@book{schmidhuber1990making,
  title={Making the world differentiable: on using self supervised fully recurrent neural networks for dynamic reinforcement learning and planning in non-stationary environments},
  author={Schmidhuber, J{\"u}rgen},
  volume={126},
  year={1990},
  publisher={Inst. f{\"u}r Informatik}
}

@article{wegner1987paradoxical,
  title={Paradoxical effects of thought suppression.},
  author={Wegner, Daniel M and Schneider, David J and Carter, Samuel R and White, Teri L},
  journal={Journal of personality and social psychology},
  volume={53},
  number={1},
  pages={5},
  year={1987},
  publisher={American Psychological Association}
}

@article{wiedemer2025video,
  title={Video models are zero-shot learners and reasoners},
  author={Wiedemer, Thadd{\"a}us and Li, Yuxuan and Vicol, Paul and Gu, Shixiang Shane and Matarese, Nick and Swersky, Kevin and Kim, Been and Jaini, Priyank and Geirhos, Robert},
  journal={arXiv preprint arXiv:2509.20328},
  year={2025}
}

@inproceedings{motamed2026generative,
  title={Do generative video models understand physical principles?},
  author={Motamed, Saman and Culp, Laura and Swersky, Kevin and Jaini, Priyank and Geirhos, Robert},
  booktitle={Proceedings of the IEEE/CVF Winter Conference on Applications of Computer Vision},
  pages={948--958},
  year={2026}
}

@inproceedings{alhamoud2025vision,
  title={Vision-language models do not understand negation},
  author={Alhamoud, Kumail and Alshammari, Shaden and Tian, Yonglong and Li, Guohao and Torr, Philip HS and Kim, Yoon and Ghassemi, Marzyeh},
  booktitle={Proceedings of the Computer Vision and Pattern Recognition Conference},
  pages={29612--29622},
  year={2025}
}

@inproceedings{parcalabescu2022valse,
  title={VALSE: A task-independent benchmark for vision and language models centered on linguistic phenomena},
  author={Parcalabescu, Letitia and Cafagna, Michele and Muradjan, Lilitta and Frank, Anette and Calixto, Iacer and Gatt, Albert},
  booktitle={Proceedings of the 60th Annual Meeting of the Association for Computational Linguistics (Volume 1: Long Papers)},
  pages={8253--8280},
  year={2022}
}

@article{conwell2024relations,
  title={Relations, negations, and numbers: Looking for logic in generative text-to-image models},
  author={Conwell, Colin and Tawiah-Quashie, Rupert and Ullman, Tomer},
  journal={arXiv preprint arXiv:2411.17066},
  year={2024}
}

@inproceedings{truong2023language,
  title={Language models are not naysayers: an analysis of language models on negation benchmarks},
  author={Truong, Thinh Hung and Baldwin, Timothy and Verspoor, Karin and Cohn, Trevor},
  booktitle={Proceedings of the 12th Joint Conference on Lexical and Computational Semantics (* SEM 2023)},
  pages={101--114},
  year={2023}
}

@inproceedings{garcia2023not,
  title={This is not a dataset: A large negation benchmark to challenge large language models},
  author={Garc{\'\i}a-Ferrero, Iker and Altuna, Bego{\~n}a and Alvez, Javier and Gonzalez-Dios, Itziar and Rigau, German},
  booktitle={Proceedings of the 2023 conference on empirical methods in natural language processing},
  pages={8596--8615},
  year={2023}
}

@article{wang2026very,
  title={A very big video reasoning suite},
  author={Wang, Maijunxian and Wang, Ruisi and Lin, Juyi and Ji, Ran and Wiedemer, Thadd{\"a}us and Gao, Qingying and Luo, Dezhi and Qian, Yaoyao and Huang, Lianyu and Hong, Zelong and others},
  journal={arXiv preprint arXiv:2602.20159},
  year={2026}
}

@article{bear2021physion,
  title={Physion: Evaluating physical prediction from vision in humans and machines},
  author={Bear, Daniel M and Wang, Elias and Mrowca, Damian and Binder, Felix J and Tung, Hsiao-Yu Fish and Pramod, RT and Holdaway, Cameron and Tao, Sirui and Smith, Kevin and Sun, Fan-Yun and others},
  journal={arXiv preprint arXiv:2106.08261},
  year={2021}
}

@article{tung2023physion++,
  title={Physion++: Evaluating physical scene understanding that requires online inference of different physical properties},
  author={Tung, Hsiao-Yu and Ding, Mingyu and Chen, Zhenfang and Bear, Daniel and Gan, Chuang and Tenenbaum, Josh and Yamins, Dan and Fan, Judith and Smith, Kevin},
  journal={Advances in Neural Information Processing Systems},
  volume={36},
  pages={67048--67068},
  year={2023}
}

@inproceedings{ates2022craft,
  title={Craft: A benchmark for causal reasoning about forces and interactions},
  author={Ates, Tayfun and Ate{\c{s}}o{\u{g}}lu, M and Yi{\u{g}}it, {\c{C}}a{\u{g}}atay and Kesen, Ilker and Kobas, Mert and Erdem, Erkut and Erdem, Aykut and Goksun, Tilbe and Yuret, Deniz},
  booktitle={Findings of the Association for Computational Linguistics: ACL 2022},
  pages={2602--2627},
  year={2022}
}

@article{riochet2018intphys,
  title={Intphys: A framework and benchmark for visual intuitive physics reasoning},
  author={Riochet, Ronan and Castro, Mario Ynocente and Bernard, Mathieu and Lerer, Adam and Fergus, Rob and Izard, V{\'e}ronique and Dupoux, Emmanuel},
  journal={arXiv preprint arXiv:1803.07616},
  year={2018}
}

@article{baradel2019cophy,
  title={Cophy: Counterfactual learning of physical dynamics},
  author={Baradel, Fabien and Neverova, Natalia and Mille, Julien and Mori, Greg and Wolf, Christian},
  journal={arXiv preprint arXiv:1909.12000},
  year={2019}
}

@article{yi2019clevrer,
  title={Clevrer: Collision events for video representation and reasoning},
  author={Yi, Kexin and Gan, Chuang and Li, Yunzhu and Kohli, Pushmeet and Wu, Jiajun and Torralba, Antonio and Tenenbaum, Joshua B},
  journal={arXiv preprint arXiv:1910.01442},
  year={2019}
}

@inproceedings{rajani2020esprit,
  title={ESPRIT: Explaining solutions to physical reasoning tasks},
  author={Rajani, Nazneen Fatema and Zhang, Rui and Tan, Yi Chern and Zheng, Stephan and Weiss, Jeremy and Vyas, Aadit and Gupta, Abhijit and Xiong, Caiming and Socher, Richard and Radev, Dragomir},
  booktitle={Proceedings of the 58th Annual Meeting of the Association for Computational Linguistics},
  pages={7906--7917},
  year={2020}
}

@article{kang2024far,
  title={How far is video generation from world model: A physical law perspective},
  author={Kang, Bingyi and Yue, Yang and Lu, Rui and Lin, Zhijie and Zhao, Yang and Wang, Kaixin and Huang, Gao and Feng, Jiashi},
  journal={arXiv preprint arXiv:2411.02385},
  year={2024}
}

@article{bakhtin2019phyre,
  title={Phyre: A new benchmark for physical reasoning},
  author={Bakhtin, Anton and van der Maaten, Laurens and Johnson, Justin and Gustafson, Laura and Girshick, Ross},
  journal={Advances in Neural Information Processing Systems},
  volume={32},
  year={2019}
}

@article{agarwal2025cosmos,
  title={Cosmos world foundation model platform for physical ai},
  author={Agarwal, Niket and Ali, Arslan and Bala, Maciej and Balaji, Yogesh and Barker, Erik and Cai, Tiffany and Chattopadhyay, Prithvijit and Chen, Yongxin and Cui, Yin and Ding, Yifan and others},
  journal={arXiv preprint arXiv:2501.03575},
  year={2025}
}

@article{teng2025magi,
  title={Magi-1: Autoregressive video generation at scale},
  author={Teng, Hansi and Jia, Hongyu and Sun, Lei and Li, Lingzhi and Li, Maolin and Tang, Mingqiu and Han, Shuai and Zhang, Tianning and Zhang, WQ and Luo, Weifeng and others},
  journal={arXiv preprint arXiv:2505.13211},
  year={2025}
}

@article{zhuang2025video,
  title={Video-gpt via next clip diffusion},
  author={Zhuang, Shaobin and Huang, Zhipeng and Zhang, Ying and Wang, Fangyikang and Fu, Canmiao and Yang, Binxin and Sun, Chong and Li, Chen and Wang, Yali},
  journal={arXiv preprint arXiv:2505.12489},
  year={2025}
}

@article{liu2025bootstrapping,
  title={Bootstrapping Physics-Grounded Video Generation through VLM-Guided Iterative Self-Refinement},
  author={Liu, Yang and Zhao, Xilin and Wen, Peisong and Dai, Siran and Huang, Qingming},
  journal={arXiv preprint arXiv:2511.20280},
  year={2025}
}

@article{lu2026phys4d,
  title={Phys4D: Fine-Grained Physics-Consistent 4D Modeling from Video Diffusion},
  author={Lu, Haoran and Wu, Shang and Zhang, Jianshu and Su, Maojiang and Ye, Guo and Xu, Chenwei and Lu, Lie and Maneriker, Pranav and Du, Fan and Li, Manling and others},
  journal={arXiv preprint arXiv:2603.03485},
  year={2026}
}

@article{agarwal2026cosmos,
  title={Cosmos 3: Omnimodal World Models for Physical AI},
  author={Agarwal, Niket and Ali, Arslan and Allen, Jon and Antolini, Martin and Aubame, Adeline and Azzolini, Alisson and Bai, Junjie and Bala, Maciej and Balaji, Yogesh and Bapst, Josh and others},
  journal={arXiv preprint arXiv:2606.02800},
  year={2026}
}

@inproceedings{xue2025phyt2v,
  title={Phyt2v: Llm-guided iterative self-refinement for physics-grounded text-to-video generation},
  author={Xue, Qiyao and Yin, Xiangyu and Yang, Boyuan and Gao, Wei},
  booktitle={Proceedings of the Computer Vision and Pattern Recognition Conference},
  pages={18826--18836},
  year={2025}
}

@InProceedings{huang2024vbench,
  title     = {{VBench}: Comprehensive Benchmark Suite for Video Generative Models},
  author    = {Huang, Ziqi and He, Yinan and Yu, Jiashuo and Zhang, Fan and
               Si, Chenyang and Jiang, Yuming and Zhang, Yuanhan and Wu, Tianxing and
               Jin, Qingyang and Chanpaisit, Nattapol and Wang, Yaohui and
               Chen, Xinyuan and Wang, Limin and Lin, Dahua and Qiao, Yu and Liu, Ziwei},
  booktitle = {Proceedings of the IEEE/CVF Conference on Computer Vision and
               Pattern Recognition (CVPR)},
  year      = {2024}
}

@article{huang2024vbenchpp,
  title   = {{VBench++}: Comprehensive and Versatile Benchmark Suite for Video Generative Models},
  author  = {Huang, Ziqi and Zhang, Fan and Xu, Xiaojie and He, Yinan and
             Yu, Jiashuo and Dong, Ziyue and Ma, Qianli and Chanpaisit, Nattapol and
             Si, Chenyang and Jiang, Yuming and Wang, Yaohui and Chen, Xinyuan and
             Chen, Ying-Cong and Wang, Limin and Lin, Dahua and Qiao, Yu and Liu, Ziwei},
  journal = {arXiv preprint arXiv:2411.13503},
  year    = {2024}
}

@article{zheng2025vbench2,
  title   = {{VBench-2.0}: Advancing Video Generation Benchmark Suite for Intrinsic Faithfulness},
  author  = {Zheng, Dian and Huang, Ziqi and Liu, Hongbo and Zou, Kai and He, Yinan and
             Zhang, Fan and Zhang, Yuanhan and He, Jingwen and Zheng, Wei-Shi and
             Qiao, Yu and Liu, Ziwei},
  journal = {arXiv preprint arXiv:2503.21755},
  year    = {2025}
}

@InProceedings{liu2024evalcrafter,
  title     = {{EvalCrafter}: Benchmarking and Evaluating Large Video Generation Models},
  author    = {Liu, Yaofang and Cun, Xiaodong and Liu, Xuebo and Wang, Xintao and
               Zhang, Yong and Chen, Haoxin and Liu, Yang and Zeng, Tieyong and
               Chan, Raymond and Shan, Ying},
  booktitle = {Proceedings of the IEEE/CVF Conference on Computer Vision and
               Pattern Recognition (CVPR)},
  year      = {2024}
}

@article{bansal2024videophy,
  title   = {{VideoPhy}: Evaluating Physical Commonsense for Video Generation},
  author  = {Bansal, Hritik and Lin, Zongyu and Xie, Tianyi and Zong, Zeshun and
             Yarom, Michal and Bitton, Yonatan and Jiang, Chenfanfu and
             Sun, Yizhou and Chang, Kai-Wei and Grover, Aditya},
  journal = {arXiv preprint arXiv:2406.03520},
  year    = {2024}
}

@article{wang2024vamp,
  title   = {What You See Is What Matters: A Novel Visual and Physics-Based Metric
             for Evaluating Video Generation Quality},
  author  = {Wang, Zihan and Li, Songlin and Hao, Lingyan and Hu, Xinyu and Song, Bowen},
  journal = {arXiv preprint arXiv:2411.13609},
  year    = {2024}
}
}

\newpage
\appendix
\renewcommand{\contentsname}{Appendix Contents} 
\setcounter{tocdepth}{2} 
\part{Appendix} 

\parttoc 
\clearpage
\section{Prompt Improvements}
\subsection{Qualitative Prompt Examples}
\begin{figure}[h]
    \centering
    \includegraphics[width=\linewidth]{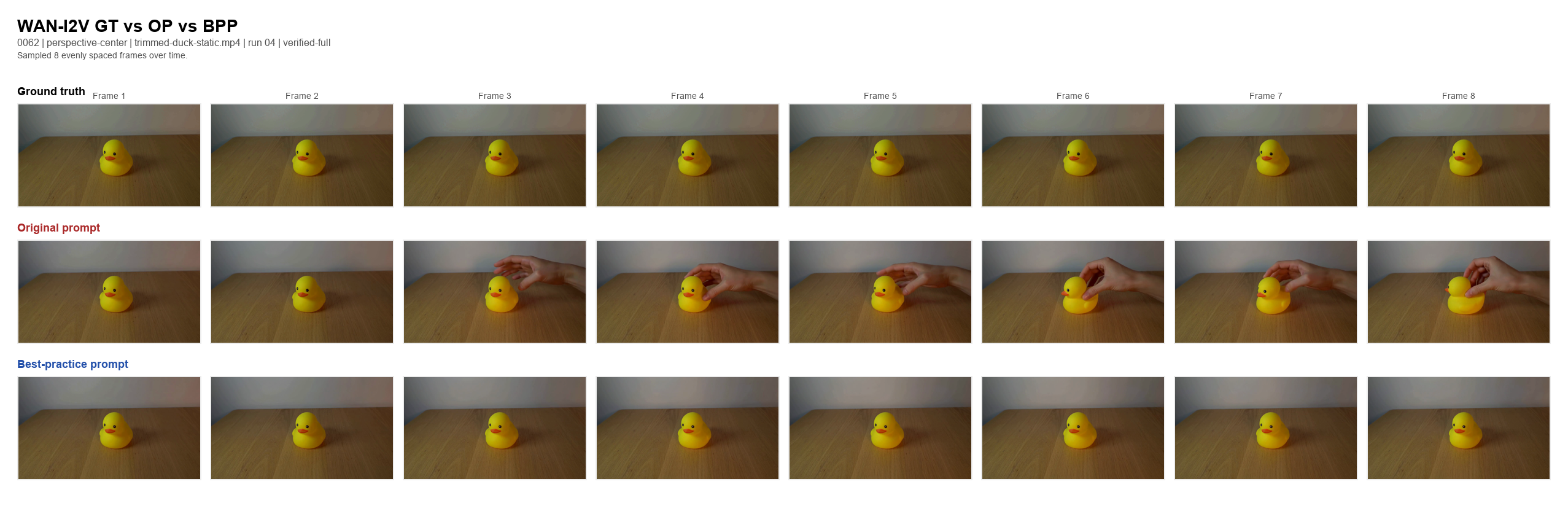}
    \caption{
    \textbf{Comparison between a generation with the original prompt and verified prompt} using Wan 2.2 to generate a static rubber duck on a wooden table. Using the original prompt a hand appears and interacts with the duck. The Best Practice Prompt has explicit description that nothing except the described phenomena occurs.
    \\
    \textbf{Original Prompt}: A stationary yellow rubber duck on a light brown wooden table against a plain white background. Static shot with no camera movement.
    \\
    \textbf{Best Practice Prompt}: The yellow rubber duck sits stationary on a light brown wooden table., Behind the wooden table is a plain white background., Static locked-off single-shot with fixed frame throughout filmed with constant framerate in real-time., The scene shows a realistic scientific demonstration., The scene only contains the described setup and actions.
    }
    \label{fig:bpp-rubberduck}
\end{figure}

\begin{figure}[h]
    \centering
    \includegraphics[width=\linewidth]{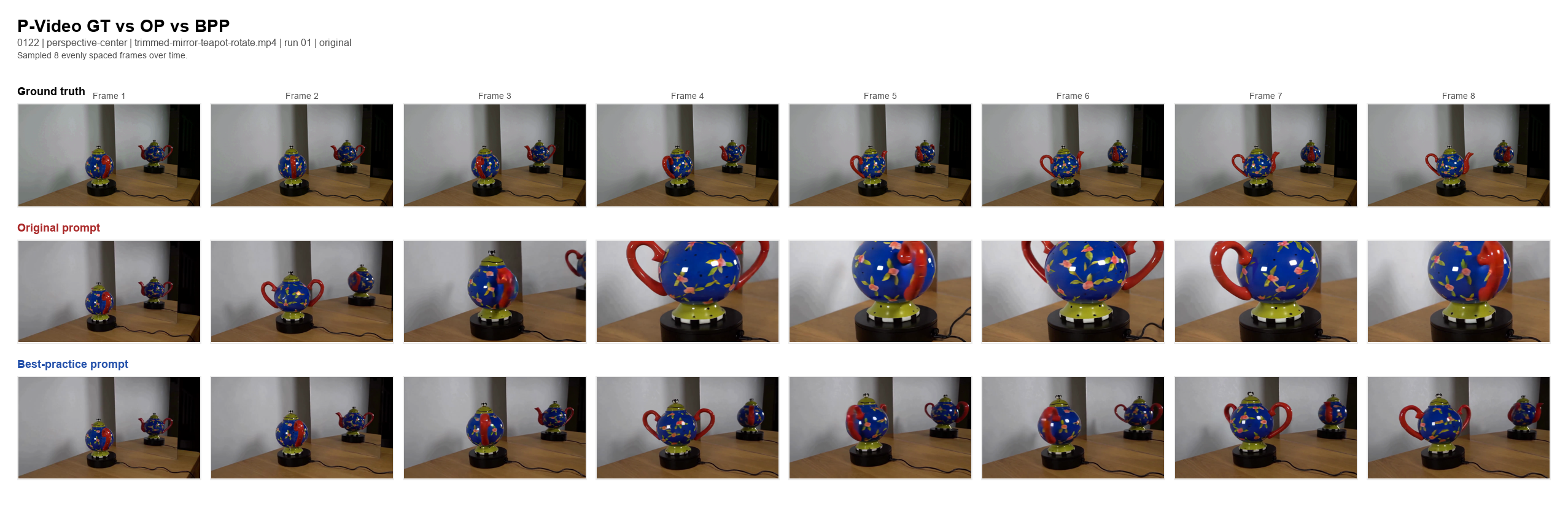}
    \caption{
    \textbf{Comparison between a generation with the original prompt and verified prompt} using p-video to generate a rotating teapot in front of a mirror. Using the original prompt the camera zooms in. The Best Practice Prompt has explicit description that the camera remains in position.
    \\
    \textbf{Original Prompt}: A teapot on a rotating display base that rotates clockwise in front of a mirror reflecting the teapot's image. Static shot with no camera movement.
    \\
    \textbf{Best Practice Prompt}: The teapot rotates clockwise on the black platform in front of a mirror that reflects the teapot's image., Static locked-off single-shot with fixed frame throughout filmed with constant framerate in real-time., The scene shows a realistic scientific demonstration., The scene only contains the described setup and actions.
    }
    \label{fig:bpp-teapot}
\end{figure}

\begin{figure}[h]
    \centering
    \includegraphics[width=\linewidth]{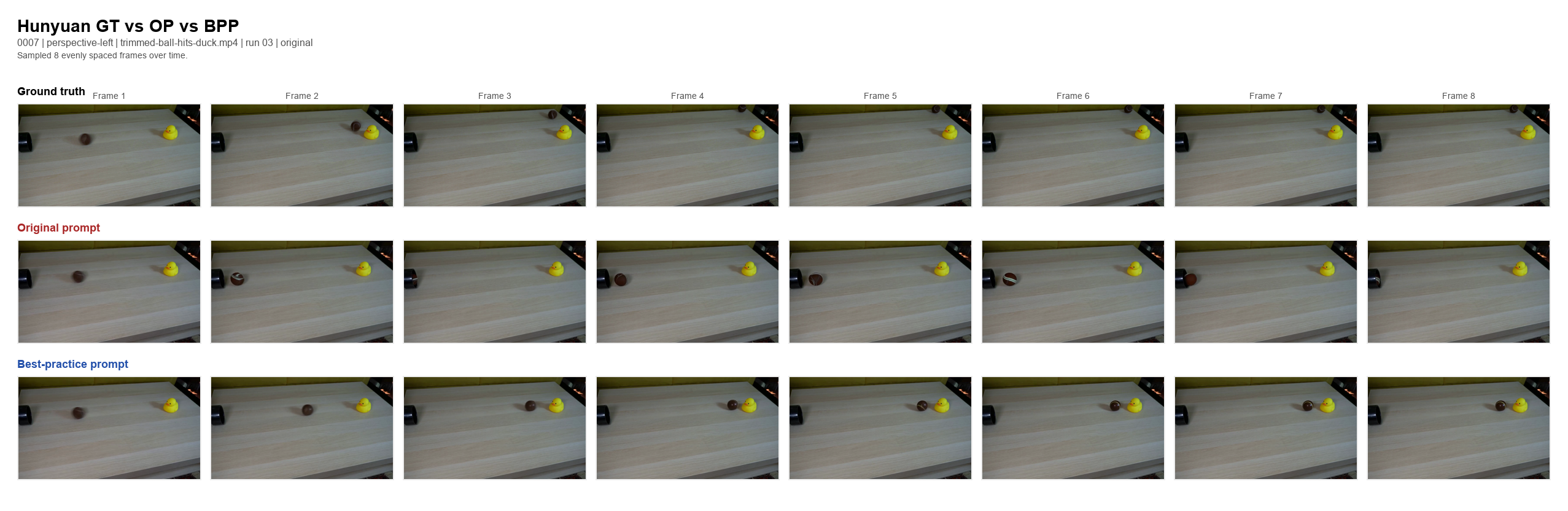}
    \caption{
    \textbf{Comparison between a generation with the original prompt and verified prompt} using  HunyuanV-1.5 to generate a tennis ball hitting a rubber duck. Using the original prompt there is no information regarding speed and the ball stops. The Best Practice Prompt has as additional information a proxy for the speed of the ball.
    \\
    \textbf{Original Prompt}: A light beige coffee table with a small yellow rubber ducky on it. A mustard yellow couch is in the background. There is a black pipe on one end of the table and a brown tennis ball rolls out of it towards the rubber ducky. Static shot with no camera movement.
    \\
    \textbf{Best Practice Prompt}: The brown tennis ball rolls straight out of the black pipe and hits the rubber duck., A light beige coffee table with a small yellow rubber duck on it. A mustard yellow couch is in the background. There is a black pipe that points from the left side to the right side of the table. , Static locked-off single-shot with fixed frame throughout filmed with constant framerate in real-time., The scene shows a realistic scientific demonstration., The scene only contains the described setup and actions.
    }
    \label{fig:bpp-ballduck}
\end{figure}

\clearpage

\subsection{Prompt Template Design}\label{app:templater}
A well-designed prompt should function as an exam question: a human given the prompt and start frame should be able to predict the experimental outcome with high confidence, but the prompt must not make the answer obvious, lest it trivialise the generation task. Any ambiguity left unresolved by the prompt introduces degrees of freedom in the output that are orthogonal to physical understanding and therefore inflate metric variance or reduces performance irreducibly.

\begin{table}[h]
\centering
\caption{Prompt template fields. The two fields marked $^*$ are novel additions not present in the original prompts. Variable fields are scenario-specific; fixed fields are shared across all 66 scenarios.}
\label{tab:prompt_fields}
\renewcommand{\arraystretch}{1.25}
\begin{tabular}{llp{7.6cm}}
\toprule
\textbf{Symbol} & \textbf{Type} & \textbf{Content} \\
\midrule
\texttt{SETUP}  & Variable & Pre-action scene description: objects, their spatial arrangement, and initial conditions prior to any physical event. \\
\texttt{SCENE}  & Variable & Scene description: supplementing \texttt{SETUP} with temporally constant information. \\
\texttt{ACTION} & Variable & Subject-action description. \\
\midrule
\texttt{CAM}    & Fixed    & Camera and recording specification; enforces a static, locked-off, constant-framerate shot. \\
\texttt{STYLE}$^*$  & Fixed    & Rendering register; constrains output to a realistic scientific demonstration. \\
\texttt{SCOPE}$^*$  & Fixed    & Content boundary; instructs the model that no new actions take place during this video. \\
\bottomrule
\end{tabular}
\end{table}

\subsection{Avoiding Negations}
A core principle of the rewrite is to express all instructions in positive terms, motivated on three independent grounds. 
From the model perspective, it is a known phenomenon that text-based negations are poorly handled which likely extends to video-models given that it has been observed for LLMs \citep{truong2023language,garcia2023not}, vision--language models such as CLIP \citep{parcalabescu2022valse,alhamoud2025vision}, and text-to-image generative models \citep{conwell2024relations} that they all exhibit systematic failures with negated instructions. 
For the human psyche, suppressing a concept reliably activates it, a phenomenon formalised as ironic process theory by \citet{wegner1987paradoxical}. 
Finally, positive framing is explicitly recommended in provider prompting guidelines.\footnote{e.g. FLUX: \url{https://docs.bfl.ml/guides/prompting_summary}}

\subsection{Camera Guidance}
Cinematographic consistency is particularly consequential for this benchmark: evaluation metrics penalise deviations in camera pose and motion between generated and ground-truth video. The original prompts specify only ``Static shot with no camera movement''.  
\citet{motamed2026generative} themselves acknowledge that given this setup many models and especially Sora are still subject to camera drift. The importance of more thorough cinematographic specification becomes clear implicitly when reading instructions like ``Static camera perspective, no zoom no pan no movement no dolly no rotation'' in \citet[Figs.~10--26]{wiedemer2025video}. 
Applying the positive-framing principle consistently, we formalise these findings into a single fixed \texttt{CAM} field: \emph{``Static locked-off single-shot with fixed frame throughout, filmed at constant framerate in real-time.''} This replaces negation-based instructions with descriptive cinematographic language and is applied uniformly across all scenarios.

\section{Artifact Cleaning and Dataset Modification}
\label{app:data_examples_improvement}
We address each artifact type through a targeted removal strategy. 
Both strategies rely on manual annotation of artifact extent, encoded as \{\text{annotation}\} in the dataset, and employ \emph{frame freezing} as the removal primitive which corresponds to holding pixel values constant in the affected region from a given timestamp onward. 
Freezing is preferred over alternatives such as masking or inpainting because it introduces no new visual information and avoids artificial boundaries that could themselves generate spurious metric activations.
\begin{itemize}[nosep, leftmargin=*]
    \item \textbf{Post-effect removal} targets artifacts occurring \emph{after} the physical effect has concluded. Frames are frozen beyond a manually annotated endpoint specified via \{\text{end\_effect\_frames}\}, eliminating all post-effect visual events regardless of their spatial location. This primarily addresses Non-deterministic artifacts in the temporal tail of the video.

    \item \textbf{Mid-effect removal} targets artifacts occurring \emph{during} the physical effect in regions that are spatially disjoint from it. Designated spatial regions are frozen from a manually annotated timestamp onward, specified via \{\text{freeze\_areas}\}. This strategy handles both deterministic apparatus artifacts and incidental non-deterministic events that overlap temporally with the effect.
\end{itemize}

We provide visual examples for representative artifact corrections and a dataset-wide overview of our applied changes in Figure~\ref{fig:actifact_tile_overview}.

\subsection{Qualitative Artifact Examples}

\begin{figure}[h]
      \centering
      \includegraphics[width=\linewidth]{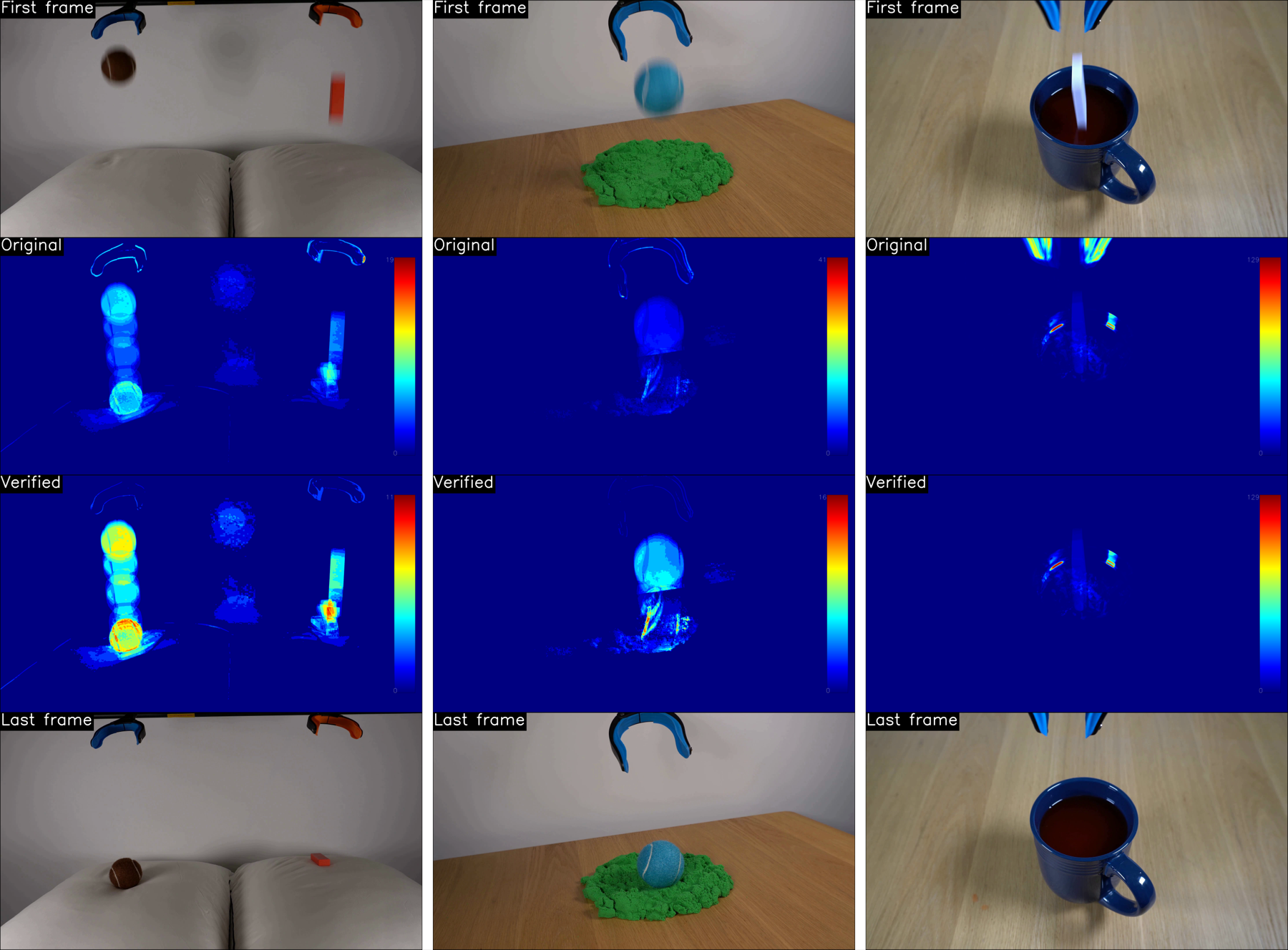}
      \caption{\textbf{Exemplary changes: Non-deterministic artifacts, here mainly grabber-related regions.} Each column shows the first frame, the original aggregated activation map, the verified aggregated activation map, and the last frame. The grabber tools glow bright in the original activation map. However, their movement is unrelated to the physical effect: the falling objects. By removing both post-effect artifacts after the objects landed and the mid-effect artifacts during the fall in a spatial region around the grabbers, the resulting activation map focuses more closely on the falling objects. }
      \label{fig:appendix_triptych_0002_0016_0047}
  \end{figure}

\begin{figure}[t]
      \centering
      \includegraphics[width=\linewidth]{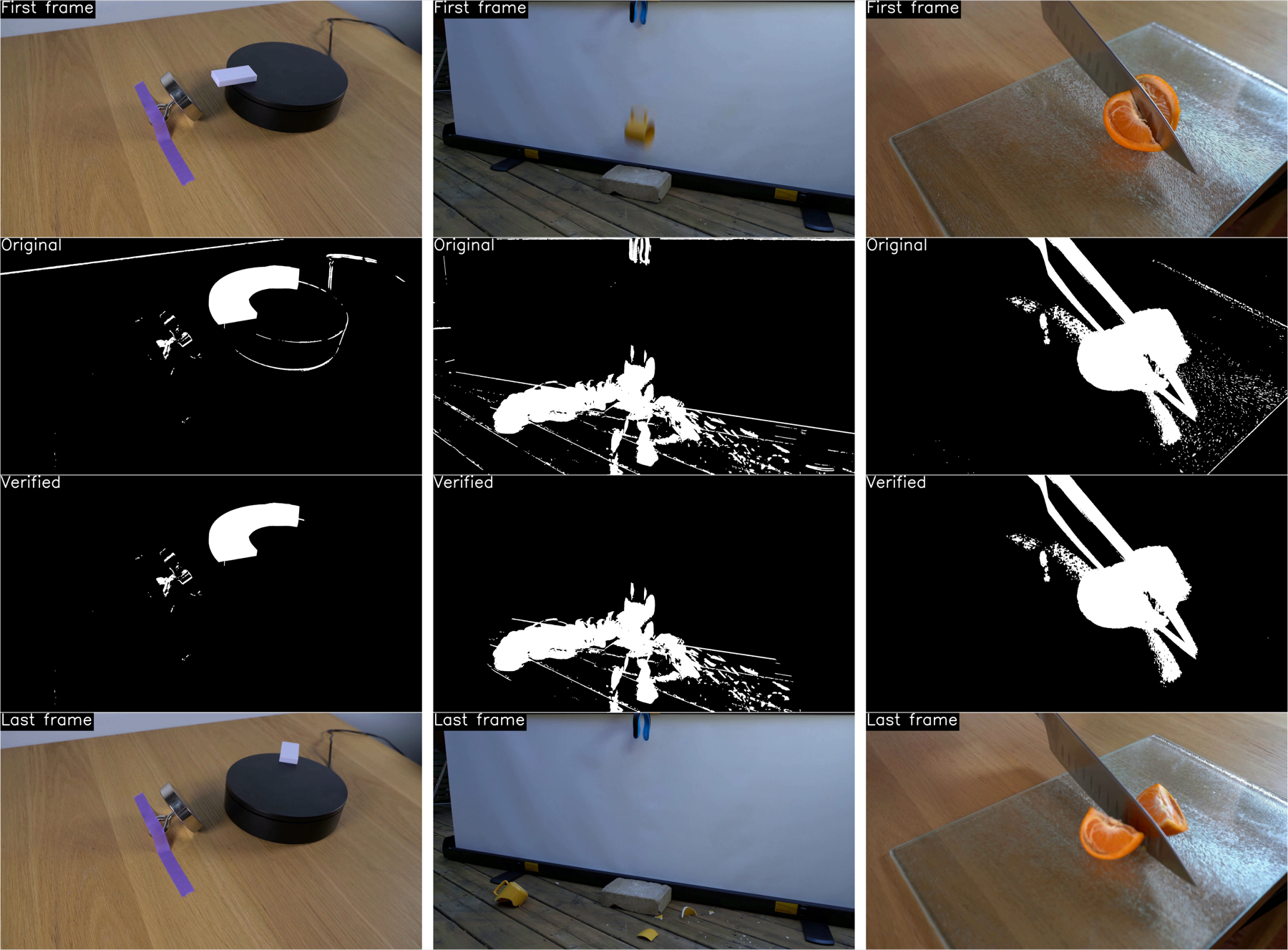}
      \caption{\textbf{Exemplary changes: Additional non-deterministic artifacts.} Each column shows the
  first frame, the original aggregated activation map, the verified aggregated activation map, and the last frame. We use binary maps here because they better reveal smaller spatial changes and make more localized random effects easier to detect. The recording errors generate activations in the binary activation map. However, their movement is unrelated to the physical effect: the (a) rotating, (b) falling or (c) object being cut. By removing both post-effect artifacts after the physical phenomena and the mid-effect artifacts during the physical phenomena, the resulting activation map focuses more closely on the falling objects.}
      \label{fig:appendix_ex_non_sys_rest_0094_0126_0040}
  \end{figure}

\begin{figure}[]
      \centering
      \includegraphics[width=\linewidth]{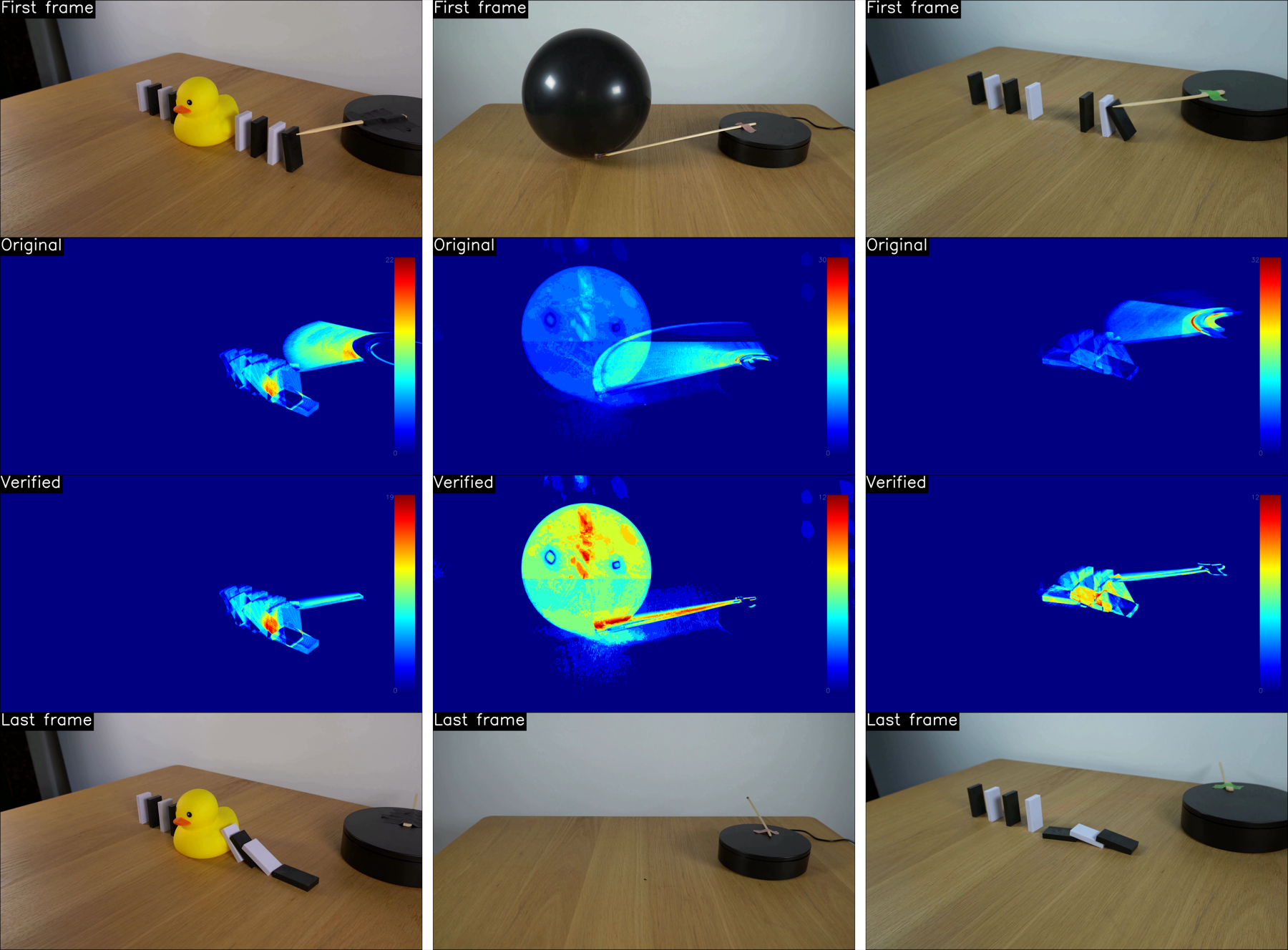}
      \caption{\textbf{Exemplary changes: Deterministic artifacts.} Each column shows the first frame, the original aggregated activation map, the verified aggregated activation map, and the last frame. Note that we modified the improved prompt in these particular cases to stop the rotating base, once the effect has been set in motion. The rotators glow bright in the original activation map. However, their movement is unrelated to the physical effect: the observed physical phenomena. By removing the effect artifacts the resulting activation map focuses more closely on the falling objects.}
      \label{fig:appendix_ex_sys_0057_0113_0051}
  \end{figure} 

\clearpage

\subsection{Dataset-Wide Modification Overview}

\begin{figure}[h]
    \centering
    \includegraphics[width=\linewidth]{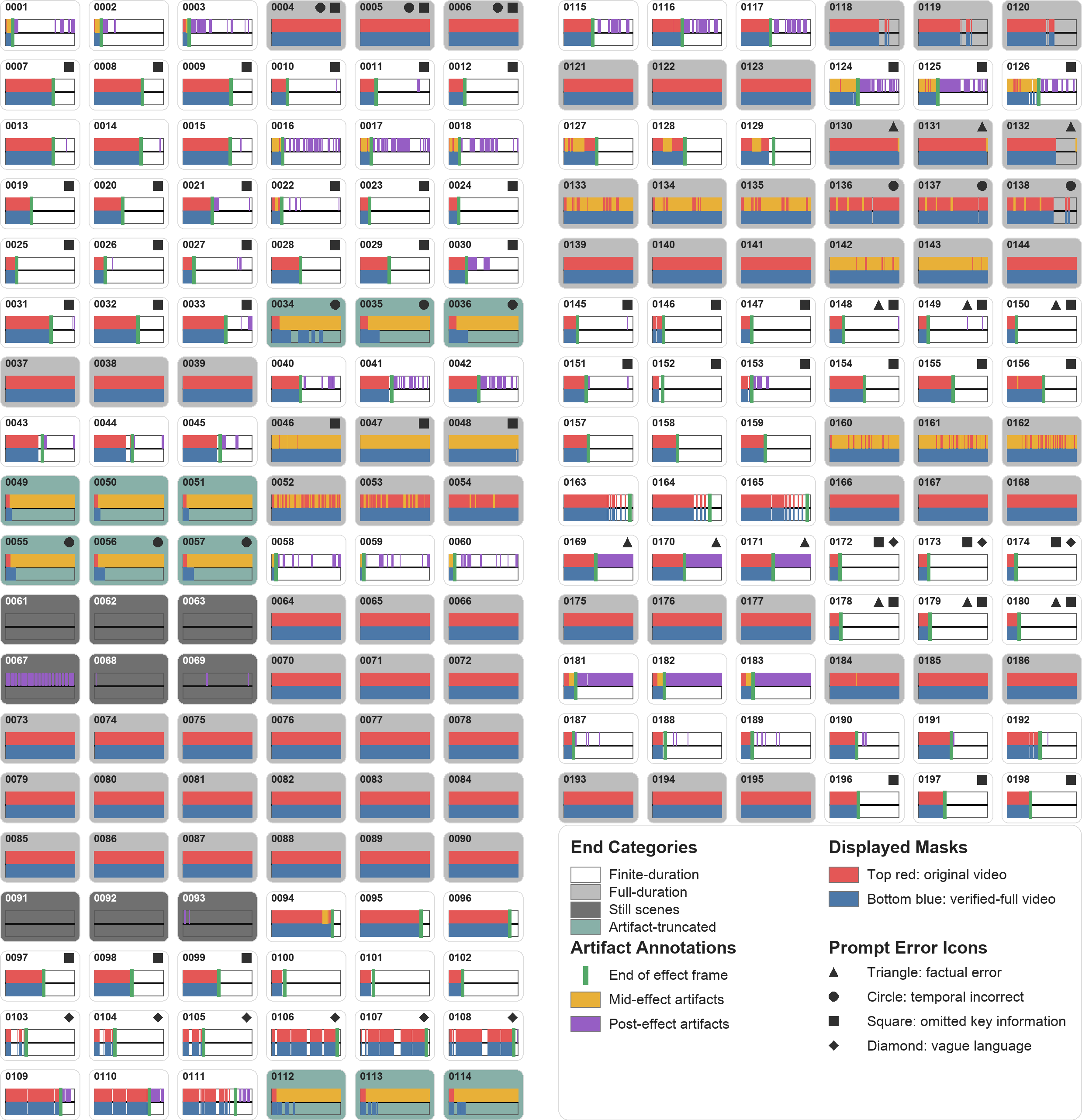}
    \caption{\textbf{Modification Overview.} \textbf{Tiles:} Each tile represents one take-1 video from the 198-video evaluation set. Red marks activity removed after the annotated effect end; blue marks activity retained in the verified evaluation; grey indicates videos whose physical effect continues throughout the full duration. The error icons mark videos, where this specific error is present in the original version.} 
    \label{fig:actifact_tile_overview}
\end{figure}

\clearpage

\section{Detailed Metric Definition} 
\label{app:metric-discussion}

\subsection{Key improvements from the original to the verified Physics-IQ evaluation.}
\label{app:phys-iq-detailed-pipeline}
\begin{figure}[h]
    \centering
    \includegraphics[width=\linewidth]{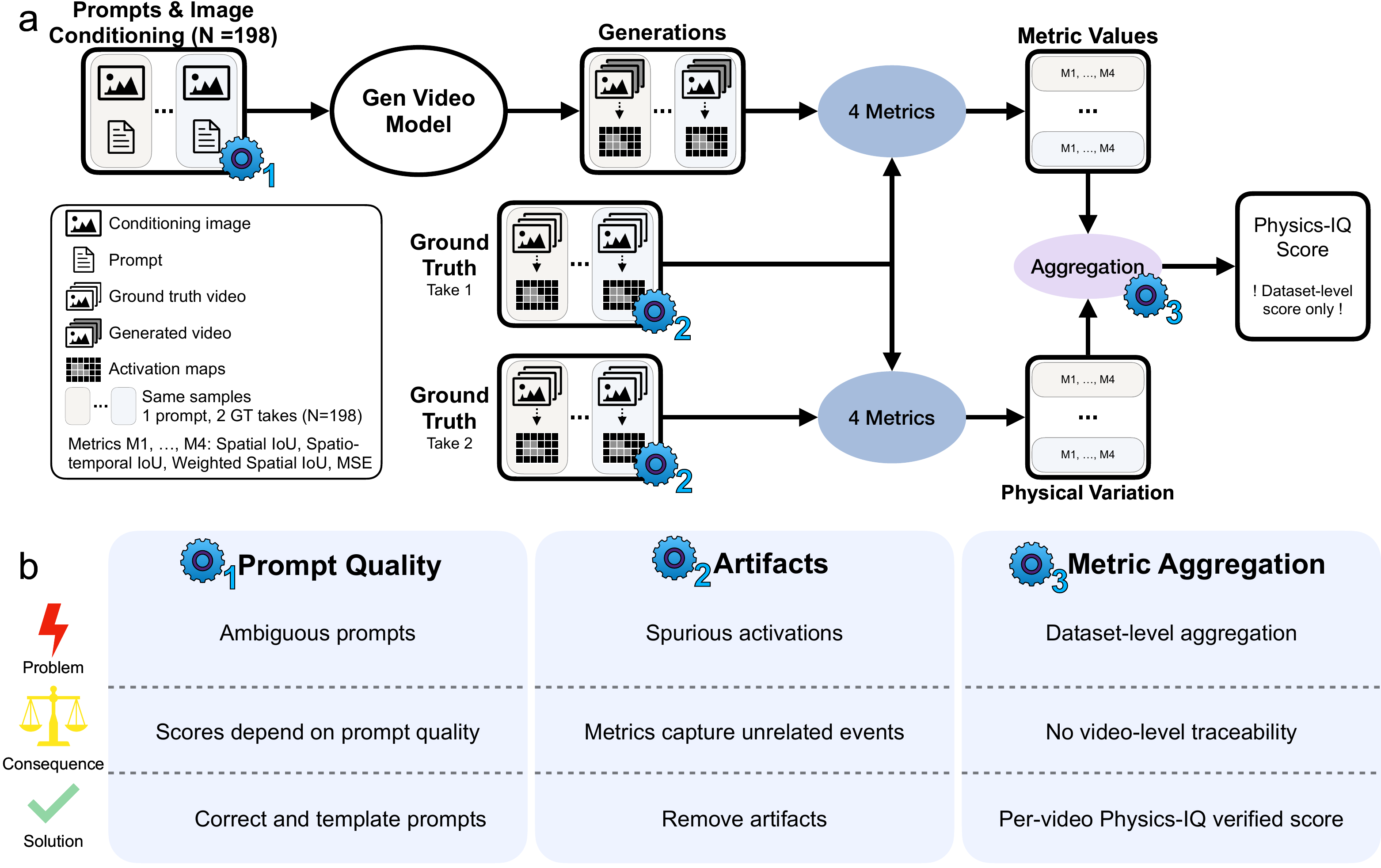}
    \caption{\textbf{Key improvements from the original to the verified Physics-IQ evaluation.} \textbf{(a)} Overview of the Physics-IQ evaluation pipeline, where a generative model produces video continuations that are compared to a ground truth using three activation-based and one pixel-based metric, followed by aggregation into a final score. Light tile colors indicate corresponding elements of the same benchmark sample: one conditioning image and prompt, one generated continuation, and two repeated ground-truth recordings (GT1 and GT2) of the same experiment. GT1 is used as the reference continuation, while GT2 is used in combination with GT1 to estimate physical variation. \textbf{(b)} We propose three refinements to the original pipeline targeting: (1) prompt quality, (2) spurious metric activations (artifacts), and (3) metric aggregation. These improvements together sharpen the focus of the evaluation on physical understanding rather than confounding factors and also lead to a fine-grained understanding of the final score in which also all samples are weighted equally.
    }
    \label{fig:main_figure_appendix_extended}
\end{figure}

\subsection{Variables and Derived Maps}
 
The dataset consists of $C \times N = E \cdot S$ videos, where $E$ experiments are each captured with $C$ takes across $S$ viewing angles. Each video is a tensor $V \in \mathbb{R}^{H \times W \times T}$, where $H \times W$ denotes spatial resolution and $T$ the number of frames.
 
From each video, a binary spatiotemporal activation map $A_{\text{source}}^{\text{ST}} \in \{0,1\}^{H \times W \times T}$ is derived from the greyscale signal, encoding where and when motion or activation occurs (see \citep[Algo. 2]{motamed2026generative} for details). Two further representations are derived from this map for use in the metrics:
\begin{itemize}
    \item The \textbf{spatial activation map} $A_{\text{source}}^{\text{SP}} \in \{0,1\}^{H \times W}$ captures \emph{where} any activation occurred across the full video.
    \item The \textbf{weighted spatial activation map}
    $A_{\text{source}}^{WS} = \tfrac{1}{T} \sum_{t=1}^{T} A_{\text{source};\,:,\,:,\,t}^{\text{ST}} \in [0,1]^{H \times W}$
    captures \emph{where and how much} activation occurred, weighted by temporal frequency. Note that whether or not normalization is applied does not affect the resulting Weighted-Spatial-IoU score.
\end{itemize}
 
\subsection{Basic Metric Definitions}
 
Three IoU-based metrics are defined over the activation maps, and one pixel-level reconstruction metric over the raw video:
\begin{align}
    \text{Spatial-IoU}
        &= \text{IoU}^{\text{SP}}(A_1^{\text{SP}}, A_2^{\text{SP}})
         = \frac{\left|A_1^{\text{SP}} \cap A_2^{\text{SP}}\right|}{\left|A_1^{\text{SP}} \cup A_2^{\text{SP}}\right|}
         \in [0,1] \\[6pt]
    \text{Spatiotemporal-IoU}
        &= \text{IoU}^{\text{ST}}(A_1^{\text{ST}}, A_2^{\text{ST}})
         = \sum_{t=1}^T \frac{1}{T} \frac{\left|A_{1,:,:,t}^{\text{ST}} \cap A_{2,:,:,t}^{\text{ST}}\right|}{\left|A_{1,:,:,t}^{\text{ST}} \cup A_{2,:,:,t}^{\text{ST}}\right|}
         \in [0,1] \\[6pt]
    \text{Weighted-Spatial-IoU}
        &= \text{IoU}^{\text{WS}}(A_1, A_2)
         = \frac{\sum_{i=1}^{HW} \min(A_{1;i}^{w},\, A_{2;i}^{w})}
                {\sum_{i=1}^{HW} \max(A_{1;i}^{w},\, A_{2;i}^{w})}
         \in [0,1] \\[6pt]
    \text{MSE}(V_1, V_2)
        &= \frac{1}{HWT}\|V_1 - V_2\|_F
         \in [0,1]
\end{align}
where the MSE is computed as the mean over all frames of an experiment, with videos normalised to $[0,1]$.

The metric values $r^M_n$ for a single sample and the corresponding physical variation $r^M_n$ are defined as:
\begin{align}
    v^{M_{\text{IoU}}}_n &= \text{IoU}^{M}(A_{\text{GT};n}^{M},\, A_{\text{Gen};n}^{M})\\
    v^{\text{MSE}}_n &= \text{MSE}(V_{\text{GT};n},\, V_{\text{Gen};n}) \\
    r^{M_{\text{IoU}}}_n & =  \text{IoU}^{M}(A_{\text{GT};n}^{M},\, A_{\text{GT2};n}^{M})\\
    r^{M_{\text{MSE}}}_n &= \text{MSE}(V_{\text{GT};n},\, V_{\text{GT2};n})
\end{align}
 
\subsection{Original Physics-IQ Score Aggregation}

Each metric $M \in \{SP,\, ST,\, WS,\, \text{MSE}\}$ is aggregated over the $N$ evaluation videos into a mean score $\mu^M$ and a physical variation ceiling $\epsilon^M$. The ceiling is computed by comparing the two ground-truth takes of each experiment, quantifying the irreducible trial-to-trial variability of the physical phenomena.

For the IoU metrics the subscores are defined as:
\begin{align}
    \mu^{M}
        &= \frac{1}{N} \sum_{n=1}^{N} \text{IoU}^{M}(A_{\text{GT};n}^{M},\, A_{\text{Gen};n}^{M})
         \in [0,1] \\[6pt]
    \epsilon^{M}
        &= \frac{1}{N} \sum_{n=1}^{N} \text{IoU}^{M}(A_{\text{GT};n}^{M},\, A_{\text{GT2};n}^{M})
         \in [0,1] \\[6pt]
    s^{M}
        &= \frac{\mu^{M}}{\epsilon^{M}}
         \in \mathbb{R}_+
         \label{eq:score_iou}
\end{align}
For MSE, lower is better, so the ceiling is subtracted rather than used as a divisor:
\begin{align}
    \mu^{\text{MSE}}
        &= \frac{1}{N} \sum_{n=1}^{N} \text{MSE}(V_{\text{GT};n},\, V_{\text{Gen};n})
         \in [0,1] \\[6pt]
    \epsilon^{\text{MSE}}
        &= \frac{1}{N} \sum_{n=1}^{N} \text{MSE}(V_{\text{GT};n},\, V_{\text{GT2};n})
         \in [0,1] \\[6pt]
    s^{\text{MSE}}
        &= \mu^{\text{MSE}} - \epsilon^{\text{MSE}}
         \in [-1,1]
         \label{eq:score_mse}
\end{align}
 
\subsection{Stable Physics-IQ Score}
For the  original composite Physics-IQ score the three IoU sub-scores are averaged and the MSE penalty is subtracted to produce a raw composite score, which is then clipped to $[0,1]$:
\begin{equation}
    s_{\text{Physics-IQ}}
    = \operatorname{c}_{[0,1]}
    \!\left(
        \frac{1}{3}(s^{\text{SP}} + s^{\text{ST}} + s^{\text{WS}}) - s^{\text{MSE}}
    \right)
    \label{eq:physiq_original_app}
\end{equation}
where each subscore is normalised by the physical variation ceiling, representing the typical deviation between independent second takes of the same experiment. 
The structural flaw in Eq.~\ref{eq:physiq_original_app} is that the scores for each metric, Spatial, Spatiotemporal and weighted spatial IoU ($s^{\text{SP}}, s^{\text{ST}}, s^{\text{WS}} \in [0, \infty)$), are unbounded: a single exceptional subscore can dominate the composite irrespective of performance on the remaining metrics, directly contradicting the design intent of \citet{motamed2026generative} that ``no metric should be assessed in isolation.'' By construction, a subscore of $1$ for the positive metrics indicates that the generated videos match the ground truth as well as a second take would; scores above $1$ indicate that estimated ceiling performance has been surpassed, which the outer $\operatorname{c}_{[0,1]}$ does not prevent from inflating the average before aggregation.

As Physics-IQ is designed to assess physical understanding relative to natural scene variability, not to reward performance beyond second-take realism. We therefore enforce a performance ceiling at the physical variation by clipping each subscore individually before aggregation resulting in the \emph{Physics-IQ stable} composite score:
\begin{equation}
    s_{\text{Physics-IQ stable}}
    = \operatorname{c}_{[0,1]}
    \!\left(
        \frac{1}{3}
        \!\left(
            \operatorname{c}_{[0,1]}(s^{\text{SP}}) +
            \operatorname{c}_{[0,1]}(s^{\text{ST}}) +
            \operatorname{c}_{[0,1]}(s^{\text{WS}})
        \right)
        - \operatorname{c}_{[0,1]}(s^{\text{MSE}})
    \right)
    \label{eq:physiq_verified}
\end{equation}
The symmetry here is one of design intent rather than mathematical range.
For $s^{\text{SP}},~s^{\text{ST}},~s^{\text{WS}}$:~$\geq 1$ indicate better-than-ceiling performance and are clipped to $1$.
For $s^{\text{MSE}}$:~$\leq 0$ indicate better-than-ceiling pixel similarity and are clipped to $0$.
The symmetry here is one of design intent rather than mathematical range. 
Per-metric clipping ensures that no individual subscore can contribute beyond its intended $\frac{1}{3}$ share of the composite, while preserving full sensitivity in the practically relevant regime of below-ceiling performance.
This correction is principled regardless of empirical impact; where it additionally affects model rankings, this reflects the degree to which the original formula was distorted by subscore dominance.\footnote{At the time of writing the highest scores for the original Physics-IQ benchmark is at $62.6\approx\tfrac{100}{3}(\tfrac{43.9}{66.4} + \tfrac{33.9}{53.2} + \tfrac{33.9}{56.9}) - (0.5-0.2)$ \citep{yuan2026inference}. Therefore video generative models are not close yet in any metric towards hitting the performance ceiling.}

\subsection{Drawbacks of the Original Score}
\begin{itemize}
    \item The score is not defined for a single sample but only over the entire dataset. 
    $\longrightarrow$
    This makes it unclear on which samples a model performs well and on which samples it does not perform well.
    \item The mean aggregation for the physical variation leads to smaller values contributing less to the overall score. 
    $\longrightarrow$
    Samples that have a smaller physical variation and in theory also smaller scores contribute less to the overall score.
    \item The original unclipped score could have an overflow for sub-score values greater than 1 (or smaller than 0 for MSE).
    $\longrightarrow$
    A single very high score can dominate the Physics-IQ score.
\end{itemize}
\begin{figure}
    \centering
    \includegraphics[width=0.5\linewidth]{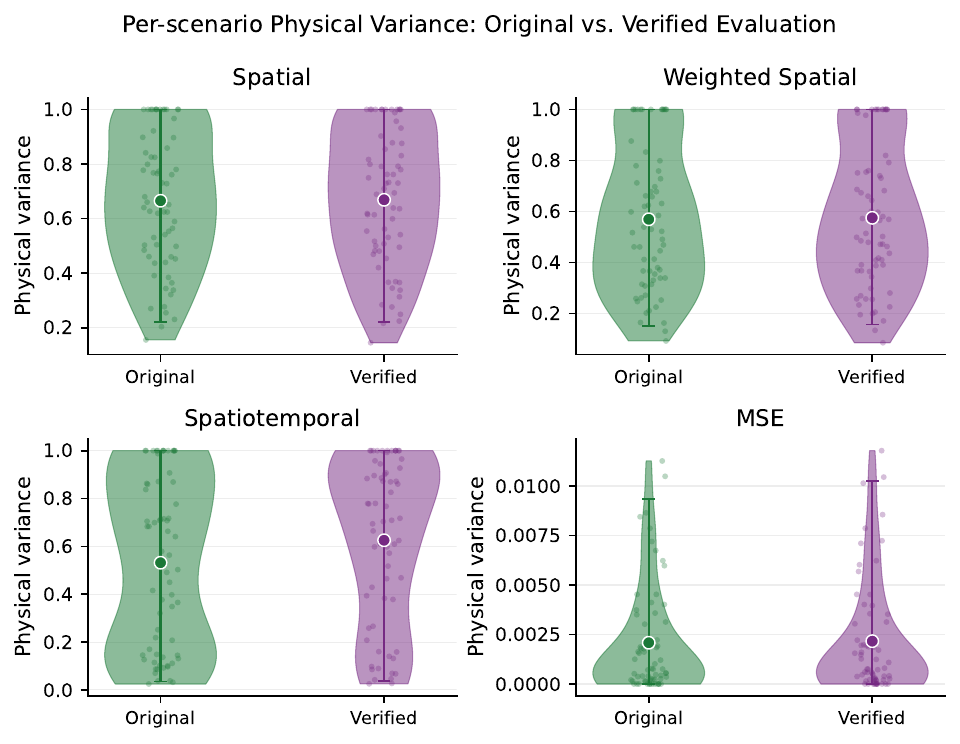}
    \caption{Visualization of the physical variance distribution $r^M_n$ per scenario obtained using the verified and the original ground truth. The results clearly indicate that the result is not gaussian distributed supporting the notion that the mean physical variance potentially downweighs the influence of samples with a low physical variance.}
    \label{fig:physvar_distribution}
\end{figure}

\subsection{Sample-Level Physics-IQ Verified Score}
We propose a principled physics-iq score operating on the sample level $i\in\{1,...,N\}$ over the entire dataset.
The aggregation of each score is performed using the 
arithmetic mean so that improvements across every single metric are clearly attributed in the per sample score.

Additionally, we change the interpretation of the MSE which now captures how many times the generated MSE score is larger than that of the physical variation.

\begin{equation}
s^{\text{Physics-IQ Verified}}_n = \frac{1}{4} \left( \underbrace{c_{[0,1]} 
\left( \frac{r^{\text{MSE}}_n}{v^{\text{MSE}}_n} \right)}_{s^{\text{MSE verified}}_n} + 
\sum_{M_{\text{IoU}} \in \{\text{SP, ST, WS}\}} 
\underbrace{
c_{[0,1]} 
\left( \frac{v^{M_{\text{IoU}}}_n}{r^{M_{\text{IoU}}}_n} \right)}
_{s^{M_{\text{IoU verified}}}_n}
\right)
\label{eq:physics-iq-verified_app}
\end{equation}
The final Physics-IQ Verified score is the arithmetic mean across 
all samples: 
\\$s^{\text{Physics-IQ Verified}} = \frac{1}{N} \sum_{n=1}^N 
s^{\text{Physics-IQ Verified}}_n$.

The subscores for each metric $s^{M\text{ verified}}$ over the entire dataset for our verified scores are obtained by summing over all samples in an identical fashion.
\clearpage
\section{Experimental Setup}

\subsection{Evaluated Models}
We provide details with respect to our evaluated models in Table~\ref{tab:model_generation_settings}.

\begin{table}[h]
\centering
  \caption{Generation settings for the evaluated image-to-video models. All models use text conditioning and a single conditioning frame. Seed control indicates whether a seed can be configured for a given model. Price via leading API providers or estimated via gpu market rate (May 2026). n.d.\ denotes values not publicly disclosed by the model provider.}
  \small
  \begin{tabular}{lcccccccc}
  \toprule
  Model & Text & v2v & i2v & Size & FPS & Resolution & Seed Control & Price\\
  \midrule
  Grok Imagine Video & $\checkmark$ & $\times$ & $\checkmark$ & n.d. & 24 & 1280$\times$720 & $\times$ & \$0.352\\
  HunyuanV-1.5 & $\checkmark$ & $\times$ & $\checkmark$ & 8.3B & 24 & 848$\times$480 & $\checkmark$ & \$0.400  \\
  P-Video & $\checkmark$ & $\times$ & $\checkmark$ & n.d. & 24 & 1280$\times$704 & $\checkmark$  & \$0.100 \\
  Sora-2 & $\checkmark$ & $\times$ & $\checkmark$ & n.d. & 30 & 1280$\times$720 & $\times$ & \$0.800\\
  Wan 2.2 & $\checkmark$ & $\times$ & $\checkmark$ & 14B & 16 & 1280$\times$720 & $\checkmark$  & \$0.110 \\
    Cosmos3-Nano & $\checkmark$ & $\times$ & $\checkmark$ & 16B & 24 & 1280$\times$720 & $\checkmark$  & \$0.333 \\
  \bottomrule
  \end{tabular}
  \label{tab:model_generation_settings}
\end{table}

For Cosmos3-Nano, we hand both op and bpp prompts directly to the \acp{VGM} without preprocessing them using a LLM or VLM.
This decision is  motivated by our aim to accurately capture the influence of the prompts, additionally the official i2v leaderboard score of Cosmos3 makes use of prompts generated using PhyT2V~\citep{xue2025phyt2v} which does not adhere to their proposed prompting structure\footnote{\href{https://huggingface.co/nvidia/Cosmos3-Nano}{recommendation for upsampling Cosmos3 Nano}, \href{https://github.com/akashgokul/cosmos/blob/feature/physicsiq-benchmark-notebook/evaluation/cosmos3/Physics_IQ/assets/i2v_prompts.json}{uploaded i2v prompts for official submission}}. Cosmos3-Nano bpp with Opus 4.8 upsampling improves over Cosmos3-Nano bpp without upsampling by \textasciitilde{}1 Physics-IQ verified score point in our separate experiments.

\section{Additional Results}
\label{app:sec_additional_results}

This section reports the full quantitative results underlying Section~\ref{s:exp}. 
We present results for both the Physics-IQ Original Score and the Physics-IQ Verified Score in Table~\ref{tab:main_results-original_score} and~\ref{tab:main_results-verified_score}.

Table~\ref{tab:sora-sanity-original_score} and~\ref{tab:sora-sanity-verified_score} provide additional Sora~2 sanity checks across evaluation dates and generation settings.

\begin{figure}[h]
    \centering
    \begin{subfigure}{0.32\linewidth}
        \includegraphics[width=\linewidth]{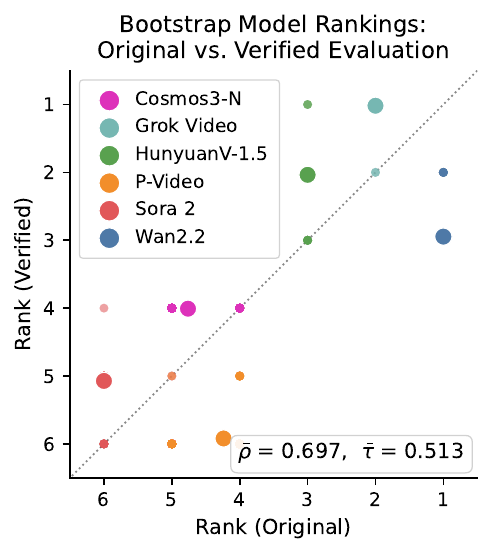}
        \caption{}
    \end{subfigure}
    \begin{subfigure}{0.48\linewidth}
        \includegraphics[width=\linewidth]{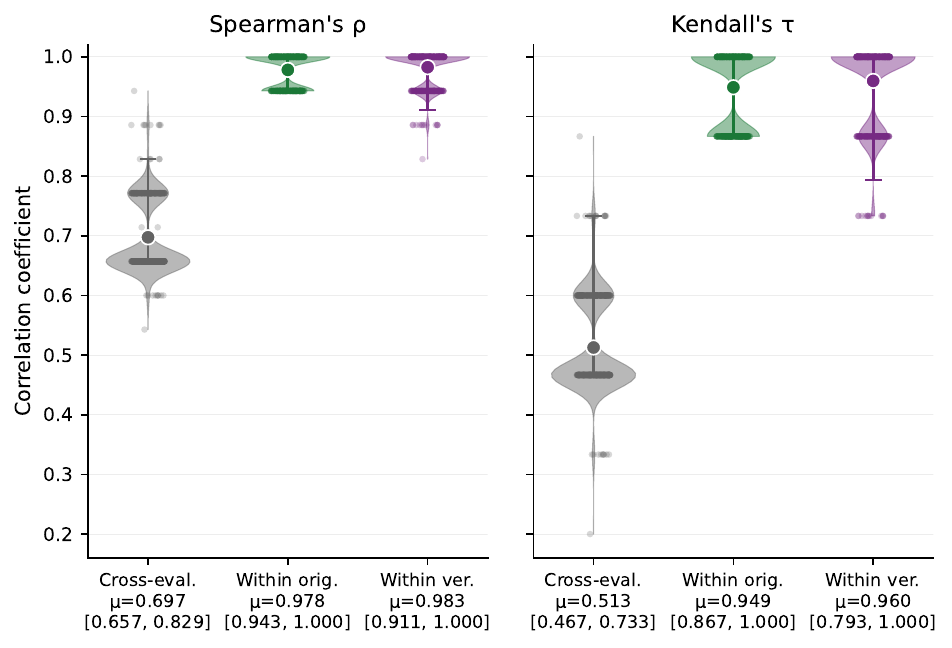}
        \caption{}
    \end{subfigure}
    \caption{\textbf{Original vs. Verified Evaluation--Ranking comparison using bootstrapping.} \textbf{(a)}~Visualization using a scatter plot, where large dots indicate the mean rank, while the smaller faint dots indicate the frequency with stronger color indicating more frequent ranks. Both the mean Spearman-$\rho$ and Kendall-$\tau$ signal meaningful ranking differences. 
    \textbf{(b)}~Distributional assessment of correlation coefficients across evaluations and within. The verified and original correlation $\tau,\rho\approx1$ indicate stable ranking within each evaluation and that the difference between both evaluations is meaningful and also outside 95\% CI intervals.
    }
    \label{fig:bootrank-analysis}
\end{figure}

\begin{figure}[h]
    \centering
    \includegraphics[width=\linewidth]{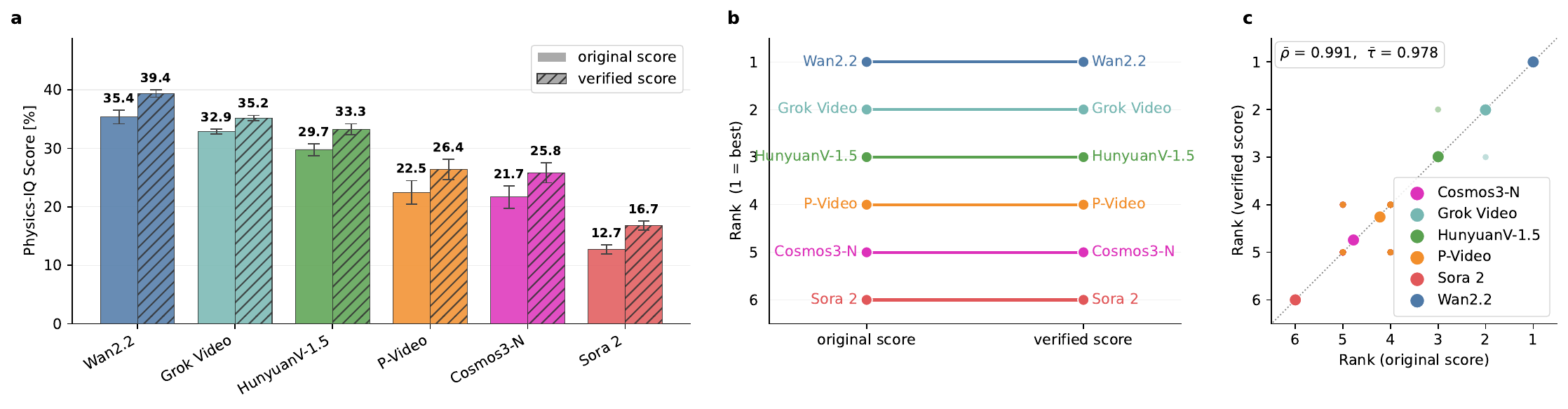}
    \caption{\textbf{Comparison of Physics-IQ scores in their original and our proposed form.} 
    \textbf{(a)}~Side-by-side comparison of original and verified Physics-IQ scores for each model. All models have higher scores.
    T-denotes the standard deviations across four different runs.
    \textbf{(b)}~Ranking bump plot showing no differences in ranking.
    \textbf{(c)}~Bootstrap analysis ranking scatter plot. Large dots indicate the mean rank, while the smaller faint dots indicate the frequency with stronger color indicating more frequent ranks. Rankings are almost perfectly aligned.
    }
    \label{fig:score-comparison}
\end{figure}

\begin{figure}[h
]
    \centering
    \begin{subfigure}{0.7\linewidth}
        \includegraphics[width=\linewidth]{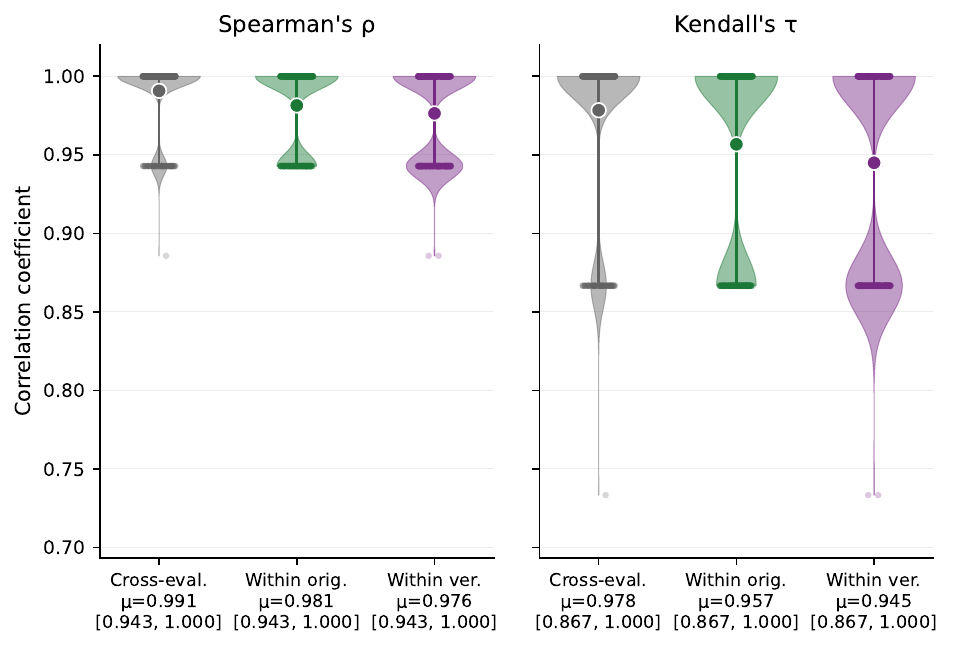}
        \caption{}
    \end{subfigure}
    \caption{\textbf{Ranking comparison using bootstrapping of Physics-IQ scores in their original and our proposed form.} \textbf{(a)}
    Distributional assessment of correlation coefficients across evaluations and within. Rankings match almost perfect.
    }
    \label{fig:score-bootstrap}
\end{figure}

Figure~\ref{fig:bootrank-analysis}, \ref{fig:score-comparison}, and \ref{fig:score-bootstrap} show the bootstrap ranking analysis used to assess ranking stability.

\clearpage
\subsection{Main Results Overview }
\label{app:subsec:main_results}
\begin{table}[h]
    \centering
    \caption{\textbf{Main Results Overview (Physics-IQ Original).} Overview of our main results. Each evaluated video-model generates four sets of videos using the original prompts (op) and the best-practice templated prompts (bpp). 
    Each of these is evaluated twice, once using the original evaluation and once using the verified evaluation where artifacts are removed. All
scores are multiplied by 100 and reported as points.
    \\
    Visualized data: $\mu \pm \sigma_{\text{STD}}$ over $4$ runs.    
    }
    \label{tab:main_results-original_score}
     \begin{adjustbox}{width=0.95\textwidth}
    \begin{tabular}{llllllll}
\toprule
 &  &  & Phys-IQ orig. &  SP orig. &  ST orig. &  WS orig. &  MSE orig. \\
Model & Ground Truth & Prompt &  &  &  &  &  \\
\midrule
\multirow[t]{4}{*}{Cosmos3-N} & \multirow[t]{2}{*}{original} & bpp & $27.8 \pm 2.2$ & $40.8 \pm 1.8$ & $19.7 \pm 3.2$ & $25.1 \pm 1.5$ & $0.8 \pm 0.1$ \\
 &  & op & $21.7 \pm 1.9$ & $33.7 \pm 1.7$ & $13.5 \pm 2.7$ & $21.4 \pm 1.2$ & $1.2 \pm 0.4$ \\
\cline{2-8}
 & \multirow[t]{2}{*}{verified} & bpp & $26.2 \pm 2.2$ & $39.8 \pm 2.1$ & $17.8 \pm 3.1$ & $23.2 \pm 1.8$ & $0.8 \pm 0.1$ \\
 &  & op & $19.7 \pm 1.9$ & $32.1 \pm 1.6$ & $11.9 \pm 2.5$ & $18.7 \pm 1.4$ & $1.2 \pm 0.4$ \\
\cline{1-8} \cline{2-8}
\multirow[t]{4}{*}{Grok Video} & \multirow[t]{2}{*}{original} & bpp & $34.8 \pm 0.8$ & $53.1 \pm 1.0$ & $15.6 \pm 0.5$ & $37.5 \pm 0.9$ & $0.6 \pm 0.0$ \\
 &  & op & $32.9 \pm 0.4$ & $50.1 \pm 0.3$ & $14.1 \pm 1.0$ & $36.2 \pm 0.3$ & $0.6 \pm 0.0$ \\
\cline{2-8}
 & \multirow[t]{2}{*}{verified} & bpp & $33.0 \pm 0.8$ & $52.2 \pm 0.9$ & $14.3 \pm 0.7$ & $34.2 \pm 0.9$ & $0.6 \pm 0.0$ \\
 &  & op & $30.6 \pm 0.7$ & $48.8 \pm 0.4$ & $12.2 \pm 1.0$ & $32.8 \pm 0.7$ & $0.6 \pm 0.0$ \\
\cline{1-8} \cline{2-8}
\multirow[t]{4}{*}{HunyuanV-1.5} & \multirow[t]{2}{*}{original} & bpp & $32.1 \pm 1.3$ & $45.4 \pm 1.7$ & $24.2 \pm 1.7$ & $28.3 \pm 1.3$ & $0.6 \pm 0.0$ \\
 &  & op & $29.7 \pm 1.0$ & $41.6 \pm 1.1$ & $23.5 \pm 1.2$ & $25.9 \pm 0.8$ & $0.6 \pm 0.0$ \\
\cline{2-8}
 & \multirow[t]{2}{*}{verified} & bpp & $31.8 \pm 1.2$ & $45.6 \pm 1.9$ & $24.1 \pm 1.3$ & $27.5 \pm 1.4$ & $0.6 \pm 0.0$ \\
 &  & op & $28.9 \pm 0.8$ & $41.0 \pm 0.9$ & $22.8 \pm 1.1$ & $24.5 \pm 0.9$ & $0.6 \pm 0.0$ \\
\cline{1-8} \cline{2-8}
\multirow[t]{4}{*}{P-Video} & \multirow[t]{2}{*}{original} & bpp & $23.7 \pm 1.8$ & $39.7 \pm 2.2$ & $11.1 \pm 2.3$ & $23.9 \pm 1.6$ & $1.2 \pm 0.1$ \\
 &  & op & $22.5 \pm 2.0$ & $36.6 \pm 1.9$ & $11.9 \pm 2.8$ & $22.7 \pm 1.4$ & $1.3 \pm 0.2$ \\
\cline{2-8}
 & \multirow[t]{2}{*}{verified} & bpp & $22.0 \pm 2.1$ & $38.2 \pm 2.2$ & $10.4 \pm 2.9$ & $20.9 \pm 1.7$ & $1.2 \pm 0.1$ \\
 &  & op & $20.6 \pm 2.1$ & $35.2 \pm 1.8$ & $10.8 \pm 3.1$ & $19.6 \pm 1.4$ & $1.3 \pm 0.2$ \\
\cline{1-8} \cline{2-8}
\multirow[t]{4}{*}{Sora 2} & \multirow[t]{2}{*}{original} & bpp & $25.3 \pm 0.8$ & $36.5 \pm 0.2$ & $22.5 \pm 2.1$ & $24.3 \pm 0.6$ & $2.5 \pm 0.1$ \\
 &  & op & $12.7 \pm 0.8$ & $23.9 \pm 0.9$ & $12.0 \pm 1.3$ & $15.2 \pm 0.5$ & $4.3 \pm 0.1$ \\
\cline{2-8}
 & \multirow[t]{2}{*}{verified} & bpp & $25.3 \pm 0.9$ & $35.6 \pm 0.3$ & $24.6 \pm 2.8$ & $23.4 \pm 0.7$ & $2.5 \pm 0.1$ \\
 &  & op & $12.4 \pm 0.9$ & $23.2 \pm 1.0$ & $12.5 \pm 1.4$ & $14.7 \pm 0.5$ & $4.4 \pm 0.1$ \\
\cline{1-8} \cline{2-8}
\multirow[t]{4}{*}{Wan2.2} & \multirow[t]{2}{*}{original} & bpp & $33.5 \pm 0.9$ & $51.9 \pm 0.9$ & $17.0 \pm 0.8$ & $33.5 \pm 1.0$ & $0.6 \pm 0.0$ \\
 &  & op & $35.4 \pm 1.2$ & $54.9 \pm 1.1$ & $17.1 \pm 1.5$ & $35.7 \pm 1.4$ & $0.5 \pm 0.0$ \\
\cline{2-8}
 & \multirow[t]{2}{*}{verified} & bpp & $29.3 \pm 0.8$ & $49.8 \pm 1.1$ & $12.9 \pm 0.6$ & $26.9 \pm 0.8$ & $0.6 \pm 0.0$ \\
 &  & op & $31.1 \pm 1.2$ & $52.6 \pm 1.1$ & $13.3 \pm 1.4$ & $28.9 \pm 1.4$ & $0.5 \pm 0.0$ \\
\cline{1-8} \cline{2-8}
\bottomrule
\end{tabular}

    \end{adjustbox}
\end{table}

\begin{table}[h]
    \centering
    \caption{\textbf{Main Results Overview (Physics-IQ Verified).} Overview of our main results. Each evaluated video-model generates four sets of videos using the original prompts (op) and the best-practice templated prompts (bpp). 
    Each of these is evaluated twice, once using the original evaluation and once using the verified evaluation where artifacts are removed. All
scores are multiplied by 100 and reported as points.
    \\
    Visualized data: $\mu \pm \sigma_{\text{STD}}$ over $4$ runs.    
    }
    \label{tab:main_results-verified_score}
     \begin{adjustbox}{width=0.95\textwidth}
    \begin{tabular}{llllllll}
\toprule
 &  &  & Phys-IQ Verified & SP verified &  ST verified &  WS verified &  MSE verified \\
Model & Ground Truth & Prompt &  &  &  &  &  \\
\midrule
\multirow[t]{4}{*}{Cosmos3-N} & \multirow[t]{2}{*}{original} & bpp & $31.2 \pm 2.5$ & $41.6 \pm 2.2$ & $25.7 \pm 3.8$ & $27.0 \pm 2.3$ & $30.6 \pm 1.9$ \\
 &  & op & $25.8 \pm 1.7$ & $35.1 \pm 2.0$ & $21.0 \pm 2.4$ & $24.0 \pm 1.3$ & $23.1 \pm 2.1$ \\
\cline{2-8}
 & \multirow[t]{2}{*}{verified} & bpp & $29.1 \pm 2.4$ & $40.4 \pm 2.3$ & $22.0 \pm 3.7$ & $24.6 \pm 2.4$ & $29.5 \pm 1.7$ \\
 &  & op & $23.3 \pm 1.6$ & $33.1 \pm 2.0$ & $17.0 \pm 2.3$ & $20.5 \pm 1.5$ & $22.5 \pm 2.0$ \\
\cline{1-8} \cline{2-8}
\multirow[t]{4}{*}{Grok Video} & \multirow[t]{2}{*}{original} & bpp & $37.3 \pm 0.6$ & $53.4 \pm 1.0$ & $25.8 \pm 0.6$ & $39.2 \pm 1.1$ & $30.8 \pm 0.4$ \\
 &  & op & $35.2 \pm 0.4$ & $50.9 \pm 0.6$ & $23.0 \pm 0.6$ & $37.4 \pm 0.3$ & $29.4 \pm 0.5$ \\
\cline{2-8}
 & \multirow[t]{2}{*}{verified} & bpp & $34.8 \pm 0.6$ & $52.7 \pm 0.9$ & $21.4 \pm 0.6$ & $35.7 \pm 1.0$ & $29.6 \pm 0.4$ \\
 &  & op & $32.7 \pm 0.4$ & $49.8 \pm 0.7$ & $18.8 \pm 0.6$ & $34.0 \pm 0.2$ & $28.2 \pm 0.4$ \\
\cline{1-8} \cline{2-8}
\multirow[t]{4}{*}{HunyuanV-1.5} & \multirow[t]{2}{*}{original} & bpp & $34.7 \pm 0.9$ & $47.5 \pm 1.1$ & $29.0 \pm 1.4$ & $31.5 \pm 0.6$ & $30.7 \pm 0.9$ \\
 &  & op & $33.3 \pm 0.9$ & $44.3 \pm 1.2$ & $28.0 \pm 0.8$ & $29.6 \pm 1.3$ & $31.2 \pm 0.5$ \\
\cline{2-8}
 & \multirow[t]{2}{*}{verified} & bpp & $33.4 \pm 0.8$ & $47.1 \pm 1.2$ & $26.9 \pm 1.0$ & $29.7 \pm 0.6$ & $30.0 \pm 1.0$ \\
 &  & op & $31.7 \pm 0.9$ & $43.5 \pm 1.1$ & $25.4 \pm 1.0$ & $27.4 \pm 1.1$ & $30.4 \pm 0.6$ \\
\cline{1-8} \cline{2-8}
\multirow[t]{4}{*}{P-Video} & \multirow[t]{2}{*}{original} & bpp & $27.6 \pm 1.8$ & $40.1 \pm 2.0$ & $19.9 \pm 2.2$ & $26.3 \pm 1.8$ & $24.3 \pm 1.3$ \\
 &  & op & $26.4 \pm 1.7$ & $37.0 \pm 1.6$ & $20.6 \pm 2.5$ & $24.9 \pm 1.4$ & $23.0 \pm 2.1$ \\
\cline{2-8}
 & \multirow[t]{2}{*}{verified} & bpp & $25.3 \pm 1.8$ & $38.6 \pm 2.2$ & $16.4 \pm 2.4$ & $22.9 \pm 1.8$ & $23.3 \pm 1.1$ \\
 &  & op & $23.8 \pm 1.7$ & $35.5 \pm 1.6$ & $16.2 \pm 2.9$ & $21.4 \pm 1.3$ & $22.2 \pm 2.0$ \\
\cline{1-8} \cline{2-8}
\multirow[t]{4}{*}{Sora 2} & \multirow[t]{2}{*}{original} & bpp & $27.3 \pm 0.8$ & $38.2 \pm 0.8$ & $28.0 \pm 1.6$ & $27.9 \pm 0.9$ & $15.0 \pm 0.6$ \\
 &  & op & $16.7 \pm 0.8$ & $24.7 \pm 1.0$ & $18.5 \pm 1.2$ & $16.3 \pm 0.5$ & $7.4 \pm 0.7$ \\
\cline{2-8}
 & \multirow[t]{2}{*}{verified} & bpp & $26.5 \pm 0.8$ & $37.3 \pm 0.6$ & $27.0 \pm 2.2$ & $26.9 \pm 0.7$ & $14.8 \pm 0.6$ \\
 &  & op & $15.7 \pm 0.7$ & $23.6 \pm 1.0$ & $16.5 \pm 1.0$ & $15.4 \pm 0.5$ & $7.4 \pm 0.6$ \\
\cline{1-8} \cline{2-8}
\multirow[t]{4}{*}{Wan2.2} & \multirow[t]{2}{*}{original} & bpp & $36.6 \pm 0.6$ & $53.2 \pm 0.7$ & $28.2 \pm 0.7$ & $35.3 \pm 0.6$ & $29.7 \pm 0.4$ \\
 &  & op & $39.4 \pm 0.6$ & $56.3 \pm 0.9$ & $29.0 \pm 0.9$ & $39.1 \pm 0.7$ & $33.1 \pm 0.3$ \\
\cline{2-8}
 & \multirow[t]{2}{*}{verified} & bpp & $32.2 \pm 0.6$ & $51.1 \pm 1.0$ & $20.5 \pm 0.7$ & $28.5 \pm 0.7$ & $28.9 \pm 0.4$ \\
 &  & op & $34.8 \pm 0.7$ & $54.3 \pm 0.9$ & $21.2 \pm 1.1$ & $31.8 \pm 0.7$ & $31.9 \pm 0.2$ \\
\cline{1-8} \cline{2-8}
\bottomrule
\end{tabular}

    \end{adjustbox}
\end{table}


\clearpage

\subsection{Sora-2 Temporal Comparison}
\label{app:subsec:sora_temporal}

\begin{table}[h]
    \centering
    \caption{\textbf{Ensuring that Sora 2 model performance is properly assessed (Physics-IQ Original).} We obtained one run in October 2025 Sora 2 (10-25) close to the original Sora 2 release which shows the highest scores, the values reported in our main paper and an additional sanity check to ensure that our generations are a valid assessment of the performance of Sora in April 2026. All scores are multiplied by 100 and reported as points. Note: For the single October run, standard deviations cannot be computed because multiple runs are required.}
    \label{tab:sora-sanity-original_score}
    \begin{adjustbox}{width=0.95\textwidth}
    \begin{tabular}{llllllll}
\toprule
 &  &  & Phys-IQ orig. &  SP orig. &  ST orig. &  weighted SP orig. &  MSE orig. \\
Model & Ground Truth & Prompt &  &  &  &  &  \\
\midrule
\multirow[t]{4}{*}{Sora 2} & \multirow[t]{2}{*}{original} & bpp & $25.3 \pm 0.8$ & $36.5 \pm 0.2$ & $22.5 \pm 2.1$ & $24.3 \pm 0.6$ & $2.5 \pm 0.1$ \\
 &  & op & $12.7 \pm 0.8$ & $23.9 \pm 0.9$ & $12.0 \pm 1.3$ & $15.2 \pm 0.5$ & $4.3 \pm 0.1$ \\
\cline{2-8}
 & \multirow[t]{2}{*}{verified} & bpp & $25.3 \pm 0.9$ & $35.6 \pm 0.3$ & $24.6 \pm 2.8$ & $23.4 \pm 0.7$ & $2.5 \pm 0.1$ \\
 &  & op & $12.4 \pm 0.9$ & $23.2 \pm 1.0$ & $12.5 \pm 1.4$ & $14.7 \pm 0.5$ & $4.4 \pm 0.1$ \\
\cline{1-8} \cline{2-8}
\multirow[t]{4}{*}{Sora 2 (12-25)} & \multirow[t]{2}{*}{original} & bpp & $24.7$ & $38.5$ & $17.5$ & $25.9$ & $2.6$ \\
 &  & op & $13.5$ & $27.0$ & $8.9$ & $17.2$ & $4.2$ \\
\cline{2-8}
 & \multirow[t]{2}{*}{verified} & bpp & $24.4$ & $37.5$ & $18.9$ & $24.9$ & $2.7$ \\
 &  & op & $12.8$ & $26.1$ & $8.6$ & $16.3$ & $4.2$ \\
\cline{1-8} \cline{2-8}
\multirow[t]{2}{*}{Sora 2 (10-25)} & original & op & $42.8$ & $55.5$ & $33.1$ & $41.4$ & $0.5$ \\
\cline{2-8}
 & verified & op & $43.6$ & $56.4$ & $33.5$ & $42.7$ & $0.5$ \\
\cline{1-8} \cline{2-8}
\bottomrule
\end{tabular}

    \end{adjustbox}
\end{table}

\begin{table}[h]
    \centering
    \caption{\textbf{Ensuring that Sora 2 model performance is properly assessed (Physics-IQ Verified).} We obtained one run in October 2025 Sora 2 (10-25) close to the original Sora 2 release which shows the highest scores, the values reported in our main paper and an additional sanity check to ensure that our generations are a valid assessment of the performance of Sora in April 2026. All scores are multiplied by 100 and reported as points. Note: For the single October run, standard deviations cannot be computed because multiple runs are required.}
    \label{tab:sora-sanity-verified_score}
    \begin{adjustbox}{width=0.95\textwidth}
    \begin{tabular}{llllllll}
\toprule
 &  &  & Phys-IQ Verified & SP verified &  ST verified &  WS verified &  MSE verified \\
Model & Ground Truth & Prompt &  &  &  &  &  \\
\midrule
\multirow[t]{4}{*}{Sora 2} & \multirow[t]{2}{*}{original} & bpp & $27.3 \pm 0.8$ & $38.2 \pm 0.8$ & $28.0 \pm 1.6$ & $27.9 \pm 0.9$ & $15.0 \pm 0.6$ \\
 &  & op & $16.7 \pm 0.8$ & $24.7 \pm 1.0$ & $18.5 \pm 1.2$ & $16.3 \pm 0.5$ & $7.4 \pm 0.7$ \\
\cline{2-8}
 & \multirow[t]{2}{*}{verified} & bpp & $26.5 \pm 0.8$ & $37.3 \pm 0.6$ & $27.0 \pm 2.2$ & $26.9 \pm 0.7$ & $14.8 \pm 0.6$ \\
 &  & op & $15.7 \pm 0.7$ & $23.6 \pm 1.0$ & $16.5 \pm 1.0$ & $15.4 \pm 0.5$ & $7.4 \pm 0.6$ \\
\cline{1-8} \cline{2-8}
\multirow[t]{4}{*}{Sora 2 (12-25)} & \multirow[t]{2}{*}{original} & bpp & $26.5$ & $41.1$ & $22.3$ & $29.6$ & $13.2$ \\
 &  & op & $17.1$ & $27.8$ & $14.8$ & $18.5$ & $7.6$ \\
\cline{2-8}
 & \multirow[t]{2}{*}{verified} & bpp & $25.8$ & $40.1$ & $21.7$ & $28.6$ & $12.8$ \\
 &  & op & $16.0$ & $26.5$ & $12.9$ & $17.3$ & $7.5$ \\
\cline{1-8} \cline{2-8}
\multirow[t]{2}{*}{Sora 2 (10-25)} & original & op & $41.1$ & $55.4$ & $37.1$ & $44.1$ & $27.9$ \\
\cline{2-8}
 & verified & op & $40.6$ & $56.0$ & $34.8$ & $44.3$ & $27.3$ \\
\cline{1-8} \cline{2-8}
\bottomrule
\end{tabular}

    \end{adjustbox}
\end{table}

\clearpage

\subsection{Bootstrap Ranking Analysis}
\label{app:subsec:bootstrap}

\clearpage

\section{Related Works}
\label{app:relworks}
\paragraph{Video generation evaluation beyond perceptual realism.} Early evaluation of video generative models (VGMs) largely focused on perceptual quality, distributional similarity, and semantic alignment, using metrics such as Fréchet Video Distance (FVD)~\cite{unterthiner2019fvd} or broad evaluation suites. More recent benchmarks decompose video quality into more fine-grained axes: VBench~\cite{huang2024vbench} and VBench++~\cite{huang2024vbenchpp} evaluate dimensions such as motion smoothness, temporal flickering, spatial consistency, subject identity, and prompt alignment, while EvalCrafter~\cite{liu2024evalcrafter} assesses visual, content, and motion quality across a diverse prompt set. VBench-2.0~\cite{zheng2025vbench2} further extends this line toward intrinsic faithfulness, including dimensions related to commonsense and physical plausibility. These benchmarks are important for measuring whether videos are visually coherent and semantically aligned, but they do not directly test whether a generated continuation follows the causal physical dynamics of a real experiment. This distinction motivates a separate line of work on physical understanding in VGMs.

\paragraph{Synthetic and simulator-based physical reasoning benchmarks.} Before the recent focus on VGMs, physical reasoning was often studied in synthetic or simulated environments. PHYRE~\cite{bakhtin2019phyre} introduced a 2D physical reasoning benchmark in which agents solve classical mechanics puzzles by interacting with a simulated world. Physion~\cite{bear2021physion} evaluates whether models can predict the future evolution of physical scenes, while Physion++~\cite{tung2023physion++} extends this setting to scenarios requiring online inference of latent physical properties. Other synthetic benchmarks, including IntPhys~\cite{riochet2018intphys}, CoPhy~\cite{baradel2019cophy}, CLEVRER~\cite{yi2019clevrer}, CRAFT~\cite{ates2022craft}, and ESPRIT~\cite{rajani2020esprit}, similarly test intuitive physics, causal reasoning, or counterfactual prediction under controlled conditions. These benchmarks provide strong experimental control and often allow large scale testing. The largest benchmark to date comprises over 10 million synthetic clips generated from 200 curated tasks, a large share of which target physical reasoning, while others cover non-physical reasoning tasks such as Sudoku \citep{wang2026very}. However, they differ from the current VGM setting because the data are typically rendered or simulated rather than recorded from real-world camera videos. Thus, they do not fully capture the visual ambiguity, apparatus effects, lighting conditions, and recording artifacts that arise when evaluating modern video generators on real physical experiments, and ultimately define the sim-to-real gap.

\paragraph{Physical reasoning benchmarks for video generative models.} Recent work has adapted physical reasoning evaluation to the VGM setting. One family relies on human or vision-language-model judgments. VideoPhy~\cite{bansal2024videophy} evaluates whether generated videos obey physical commonsense in everyday material interactions, while PhyGenBench~\cite{meng2024towards} curates prompts covering multiple physical laws and uses a hierarchical evaluation protocol. These benchmarks are scalable and cover many physical concepts, but their judgments are primarily categorical: they can identify that a generation violates a physical expectation, but they do not necessarily quantify how strongly or where the violation occurs.

A second family uses motion-, mask-, or trajectory-based proxies. VAMP~\cite{wang2024vamp} proposes visual appearance and motion-plausibility metrics based on quantities such as acceleration and velocity variance. Kang et al.~\cite{kang2024far} evaluate video generation from a physical-law perspective in synthetic environments, studying whether scaling improves the ability of VGMs to model classical mechanics. These approaches move beyond pure perceptual realism, but they either remain tied to synthetic environments or use proxy motion statistics rather than real-world reference experiments.

The third family grounds evaluation in controlled physical settings: Morpheus~\cite{zhang2025morpheus} introduces physics-informed neural networks (PINNs) to assess whether generated trajectories conform to governing equations and conserved physical invariants, such as total energy and angular momentum, derived from real laboratory experiments. This provides a complementary law-based perspective to Physics-IQ. However, Morpheus is limited to object-centric phenomena for which reliable trajectories can be extracted and for which the relevant dynamics can be expressed through low-dimensional state variables. As a result, many physical effects covered by Physics-IQ, such as drops falling into water, fluid motion, splashes, diffuse material changes, or phenomena where the relevant signal is not a single object trajectory, are not naturally captured by Morpheus-style trajectory metrics.

\paragraph{Physics-IQ.}
Physics-IQ~\cite{motamed2026generative} is the most direct predecessor of our work. It introduces a dataset of 396 real-world videos spanning five physical domains: fluid dynamics, optics, solid mechanics, magnetism, and thermodynamics.  The benchmark adopts a prediction-from-context paradigm: models are conditioned on a starting image or video clip and must generate the physical continuation of the scene. Physical understanding is evaluated through pixel-level comparison between generated and ground-truth continuations using Spatial IoU, Spatiotemporal IoU, and Mean Squared Error (MSE), aggregated into a composite Physics-IQ score normalized by the natural variance observed across real-world reference videos. Evaluations of Sora, Runway, Pika, Lumiere, Stable Video Diffusion, and VideoPoet reveal that physical understanding is severely limited across all tested models, and that visual realism is largely independent of physical accuracy~\cite{motamed2026generative} . Physics-IQ established an important empirical foundation and a reproducible evaluation pipeline. Nonetheless, several limitations remain. Mask-based overlap metrics are bounded by the quality of segmentation and may conflate spatial proximity with physical correctness. They also presuppose that the reference trajectory represents the uniquely correct physical outcome, which can yield false negatives when a generated video is physically plausible but explores a different, yet valid, realization of the scene. The benchmark does not assess whether conserved quantities such as energy or momentum are preserved in generated sequences, and the set of models evaluated has been substantially superseded by newer architectures.

\paragraph{Adoption of Physics-IQ as an benchmark and development target.} Physics-IQ has rapidly become more than a standalone benchmark. It has been used as an evaluation protocol for recent video generation systems and physics-aware model development. MAGI-1~\cite{teng2025magi} reports Physics-IQ results to assess physical continuation quality in autoregressive video generation. Yuan et al.~\cite{yuan2025improving,yuan2026inference} use Physics-IQ to evaluate whether VJEPA-2-based reward signals and inference-time alignment can improve the physical plausibility of generated videos. The ICCV 2025 Physics-IQ Challenge further institutionalized the benchmark as a shared evaluation target, with follow-up methods such as VLM-guided iterative self-refinement~\cite{liu2025bootstrapping} directly optimizing performance on the Physics-IQ task. Additional recent model papers, including Sora 2\cite{OpenAI2025Sora2}, also report Physics-IQ scores when claiming improvements in physically consistent video generation~\cite{teng2025magi,OpenAI2025Sora2,zhuang2025video,liu2025bootstrapping,lu2026phys4d}.

This adoption strengthens the motivation for our audit. Once a benchmark becomes a standard reporting protocol and an optimization target, measurement errors can propagate into model-development decisions. Prompt ambiguities, spurious ground-truth activations, and aggregation artifacts no longer only affect one benchmark paper; they can shape which systems appear more physically capable and which design choices are rewarded. Physics-IQ Verified addresses this issue by preserving the real-world continuation setting of Physics-IQ while improving prompt quality, cleaning artifact-driven activations, and introducing a sample-level aggregation scheme that makes benchmark outcomes more traceable and reliable.


\newpage
\section*{NeurIPS Paper Checklist}

\begin{enumerate}

\item {\bf Claims}
    \item[] Question: Do the main claims made in the abstract and introduction accurately reflect the paper's contributions and scope?
    \item[] Answer: \answerYes{} 
    \item[] Justification: The abstract and introduction state that the paper audits Physics-IQ and proposes three refinements: prompt improvements, artifact cleaning, and sample-level score aggregation. The claims are supported by dataset statistics and experiments on six image-to-video models.


\item {\bf Limitations}
    \item[] Question: Does the paper discuss the limitations of the work performed by the authors?
    \item[] Answer: \answerYes{} 
    \item[] Justification: We discuss limitations related to the benchmark scope, the use of a fixed set of 198 evaluation videos, dependence on manual artifact annotations, and evaluation on six image-to-video models. We also note that physically plausible but different continuations may still be penalized by reference-based evaluation.


\item {\bf Theory assumptions and proofs}
    \item[] Question: For each theoretical result, does the paper provide the full set of assumptions and a complete (and correct) proof?
    \item[] Answer: \answerNA{} 
    \item[] Justification: The paper does not introduce theoretical results or formal theorems. The mathematical content consists of metric and score definitions, which are provided in the main text and Appendix C.

    \item {\bf Experimental result reproducibility}
    \item[] Question: Does the paper fully disclose all the information needed to reproduce the main experimental results of the paper to the extent that it affects the main claims and/or conclusions of the paper (regardless of whether the code and data are provided or not)?
    \item[] Answer: \answerYes{} 
    \item[] Justification: The paper describes the dataset, evaluated models, prompt settings, generation protocol, evaluation variants, and statistical analysis needed to reproduce the main claims. Additional metric definitions and result tables are provided in the appendix.

\item {\bf Open access to data and code}
    \item[] Question: Does the paper provide open access to the data and code, with sufficient instructions to faithfully reproduce the main experimental results, as described in supplemental material?
    \item[] Answer: \answerYes{} 
    \item[] Justification: We will release the verified prompts, artifact annotations, evaluation code, and instructions for reproducing the benchmark results. The release will include anonymized access during review and full public access prior to publication (most likely within 30 days).

\item {\bf Experimental setting/details}
    \item[] Question: Does the paper specify all the training and test details (e.g., data splits, hyperparameters, how they were chosen, type of optimizer) necessary to understand the results?
    \item[] Answer: \answerYes{} 
    \item[] Justification: The experimental setup specifies the evaluated models, number of runs, prompt conditions, ground-truth variants, scoring variants, and evaluation design. Additional model and result details are provided in Appendix E and F.

\item {\bf Experiment statistical significance}
    \item[] Question: Does the paper report error bars suitably and correctly defined or other appropriate information about the statistical significance of the experiments?
    \item[] Answer: \answerYes{} 
    \item[] Justification: The paper reports standard deviations across four runs, bootstrap confidence intervals for rank correlations, and Wilcoxon signed-rank tests with Cohen’s d effect sizes for score changes.


\item {\bf Experiments compute resources}
    \item[] Question: For each experiment, does the paper provide sufficient information on the computer resources (type of compute workers, memory, time of execution) needed to reproduce the experiments?
    \item[] Answer: \answerYes{} 
    \item[] Justification: We report all model settings in the appendix. Since we just run inference and do not train, these compute resources are rather small.

    
\item {\bf Code of ethics}
    \item[] Question: Does the research conducted in the paper conform, in every respect, with the NeurIPS Code of Ethics \url{https://neurips.cc/public/EthicsGuidelines}?
    \item[] Answer: \answerYes{} 
    \item[] Justification: The research conforms to the NeurIPS Code of Ethics. It evaluates existing video generation systems on controlled benchmark data and does not involve unsafe data collection or deployment.

\item {\bf Broader impacts}
    \item[] Question: Does the paper discuss both potential positive societal impacts and negative societal impacts of the work performed?
    \item[] Answer: \answerYes{} 
    \item[] Justification: The paper discusses that more reliable physical-understanding benchmarks can improve the development and evaluation of video generative models. Potential negative impacts include strengthening video generation systems that could later be misused, although this work does not release a new generative model.
    
\item {\bf Safeguards}
    \item[] Question: Does the paper describe safeguards that have been put in place for responsible release of data or models that have a high risk for misuse (e.g., pre-trained language models, image generators, or scraped datasets)?
    \item[] Answer: \answerNA{} 
    \item[] Justification: The paper does not release a new high-risk generative model, scraped dataset, or system intended for deployment. The released assets are benchmark annotations, prompts, and evaluation code.


\item {\bf Licenses for existing assets}
    \item[] Question: Are the creators or original owners of assets (e.g., code, data, models), used in the paper, properly credited and are the license and terms of use explicitly mentioned and properly respected?
    \item[] Answer: \answerYes{} 
    \item[] Justification: The paper cites the original Physics-IQ benchmark and all evaluated models or systems. We will include license and access information for the benchmark, code dependencies, and model APIs where available.

\item {\bf New assets}
    \item[] Question: Are new assets introduced in the paper well documented and is the documentation provided alongside the assets?
    \item[] Answer: \answerYes{} 
    \item[] Justification: The paper introduces new verified prompts, artifact annotations, and evaluation code. These assets will be documented with usage instructions, data format descriptions, and limitations.


\item {\bf Crowdsourcing and research with human subjects}
    \item[] Question: For crowdsourcing experiments and research with human subjects, does the paper include the full text of instructions given to participants and screenshots, if applicable, as well as details about compensation (if any)? 
    \item[] Answer: \answerNA{} 
    \item[] Justification: The main paper does not rely on crowdsourcing or human-subject experiments. Manual artifact and prompt review was performed by the authors as part of benchmark curation.

\item {\bf Institutional review board (IRB) approvals or equivalent for research with human subjects}
    \item[] Question: Does the paper describe potential risks incurred by study participants, whether such risks were disclosed to the subjects, and whether Institutional Review Board (IRB) approvals (or an equivalent approval/review based on the requirements of your country or institution) were obtained?
    \item[] Answer: \answerNA{} 
    \item[] Justification: The paper does not involve human-subject research or crowdsourcing experiments requiring IRB or equivalent approval.

\item {\bf Declaration of LLM usage}
    \item[] Question: Does the paper describe the usage of LLMs if it is an important, original, or non-standard component of the core methods in this research? Note that if the LLM is used only for writing, editing, or formatting purposes and does \emph{not} impact the core methodology, scientific rigor, or originality of the research, declaration is not required.
    \item[] Answer: \answerNA{} 
    \item[] Justification: LLMs are not used as an important, original, or non-standard component of the core methodology. Any use for writing, editing, or formatting does not affect the scientific method or results.

\end{enumerate}

\end{document}